A Deeper Look at the Unsupervised Learning of Disentangled Representations in β-VAE from

the Perspective of Core Object Recognition

Harshvardhan Digvijay Sikka

A Thesis in the Field of Biology

for the Degree of Master of Liberal Arts in Extension Studies

Harvard University

May 2020


Abstract

The ability to recognize objects despite there being differences in appearance, known as Core Object Recognition, forms a critical part of human perception. While it is understood that the brain accomplishes Core Object Recognition through feedforward, hierarchical computations through the visual stream, the underlying algorithms that allow for invariant representations to form downstream is still not well understood. (DiCarlo et al., 2012) Various computational perceptual models have been built to attempt and tackle the object identification task in an artificial perceptual setting. Artificial Neural Networks, computational graphs consisting of weighted edges and mathematical operations at vertices, are loosely inspired by neural networks in the brain and have proven effective at various visual perceptual tasks, including object characterization and identification. (Pinto et al., 2008) (DiCarlo et al., 2012) Artificial perceptual systems often stumble when encountering the core invariance problem identifying the same object over a spectrum of transformations and viewing conditions. A popular research direction in the field of Machine Learning that attempts to solve this as a subset of a larger problem is introducing inductive biases into the model itself to reflect the structure of the input data.

The specific research problem being explored in this thesis centers on a meaningful, bounded subset of these overarching goals. For many data analysis tasks, learning representations where each dimension is statistically independent and thus disentangled from the others is useful. If the underlying generative factors of the data are


also statistically independent, Bayesian inference of latent variables can form disentangled representations. This thesis constitutes a research project exploring a generalization of the Variational Autoencoder (VAE), $\beta$-VAE, that aims to learn disentangled representations using variational inference. $\beta$-VAE incorporates the hyperparameter $\beta$, and enforces conditional independence of its bottleneck neurons, which is in general not compatible with the statistical independence of latent variables. This text examines this architecture, and provides analytical and numerical arguments, with the goal of demonstrating that this incompatibility leads to a non-monotonic inference performance in $\beta$-VAE with a finite optimal $\beta$.

Building artificial neural networks that can effectively disentangle representations is of great interest to both the neuroscience and computational perception communities. (Goodfellow et al., 2016) (LeCun et al., 2015) For the former, these models can inform scientific understanding of how neurons in the Ventral Visual Stream may be disentangling representations of objects to ascertain their identity in real time, and these systems can also provide powerful analytical tools for neuroscientists and computational biologists to apply to their own data, disentangling representations in underlying neural data to make better sense of what neuronal populations are doing. (Gurtubay et al., 2015) For the computational perception and machine learning community, building better artificial neural networks that can disentangle representations provides a powerful foundation towards more effective perceptual systems being used in modern technology. (Geron et al., 2017) Downstream impact includes better unsupervised preprocessing for

semi supervised networks and applications in various industries including transportation,

commerce, and security. (Goodfellow et al., 2016)

Acknowledgments

I wish to express my deepest gratitude to everyone whose assistance was indispensable in the completion of this project and my master's degree: Dr. Cengiz Pehlevan, Mr. Weishun Zhong, and Dr. James Morris. Your guidance, insight and support has been invaluable to me throughout my nascent research career. I would also like to thank my parents Digvijay and Nidhi Sikka, without whom none of my endeavors thus far would have been possible.



Table of Contents







Chapter I.

Introduction

Recognizing various objects in the environment makes up a fundamental part of the human perceptual experience, a task so natural and efficient that it hardly seems noteworthy. Humans have been observed classifying specific objects from numerous potential candidates in minute timescales on the order of a fifth of a second (Gurtubay et al., 2015). Despite its normalcy in our everyday experiences, the underlying computational and mechanistic processes that facilitate object recognition in the brain are far from simple. The brain itself has large portions dedicated solely to visual processing (Van Essen et al., 1992), and this aligns with the necessity for core object recognition in our daily lives, including our survival in a complex environment. As a result of this understanding, there has been a marked interest in the past decade in understanding how the brain efficiently solves the need to rapidly recognize objects despite the substantial noise and differences in their configuration and appearance (Pinto et al., 2011). From an engineering perspective, the brain presents itself as a robust construct that has efficiently solved this problem and may hold numerous useful computational paradigms that will allow for the efficient solution of the core object recognition problem in artificial systems (DiCarlo et al., 2012) This interest sits broadly at the intersection of several fields of research and study, namely neuroscience, physiology, electrical engineering, machine learning, computervision, and others. Among the various fields in which object recognition is discussed, there are numerous levels of abstraction at which it is analyzed,



ranging from specific mechanistic interactions between individual cells to broader object recognition being observed at the population or even full brain level. It should be noted that the brain solves several related but distinct problems making use of the other senses and what are surely numerous interesting algorithms and principles from biocomputation (Rolls, 2012), but the focus of this research is the investigation of core object recognition, specifically the perspective of disentangling representations of objects in an environment in Artificial Neural Network. (DiCarlo et al., 2012)

Core Object Recognition in the Brain

The concept of core object recognition is often intertwined with numerous other visual tasks, and accordingly has been scoped down to a bounded definition by researchers in the field, namely the ability to assign categorical identities to specific objects or other similar phenomena in the environment. This broad definition is being considered formally in this thesis, and we consider it more specifically as solving the object recognition objective despite various transformations of the object in the environment, including translations, rotations, and others, known more generally as object invariance (Pinto et al., 2008) (DiCarlo et al., 2012). This definition is not unique to the work outlined in this thesis, and is commonplace because its simplicity allows for meaningful analysis on tractable phenomena (Higgins et al., 2018). The object recognition problem is of particular interest in artificial systems because of the addition of the notion of invariance, as disentangling latent representations has proven to be a



difficult task for modern computer vision systems focused on object recognition (Ullman, 2000) (Pinto et al., 2008). In a real world setting, objects are perceived in an enormous range of potential images that all have the same goal classification, i.e. a coffee mug should be a coffee mug regardless of its orientation or how the environment affects its position. There are numerous variables that come into play, including but not limited to scale, pose, position, illumination, other objects, and obscuration. The task becomes even more challenging when one considers that many objects do not maintain the same structure in all situations. (Higgins et al., 2018) These problems also exist in different forms at the individual identification level up to the categorical identification level, i.e. what constitutes a car? Accordingly, there are countless potential activations that can occur in response to the emerging orientations as a result of all the possible variables affecting a 3 dimensional object, and the brain robustly makes equivalent connections between these activations, in fractions of a second. (Gurtubay et al., 2015) (Higgins et al., 2018)

In primates, the neural circuits that have a role in core object recognition are a part of the Ventral Visual Stream, and with the inferior temporal cortex (IT). The Ventral Visual stream can be further broken down into hierarchically organized areas. Information is processed in the Ventral Visual Stream through neuronal spiking and signal propagation along axons, and areas are characterized by representations of spiking patterns in their neuronal population. (Hung et al., 2005) The processing in these populations of neurons in downstream areas, including IT, is still being understood, though it has been observed that IT patterns correspond with real time invariant object



characterization (Li et al., 2009). These populations are also better at the object categorization task than earlier areas like V1 or V2. The IT population also begins to demonstrate that visual object information is made available around 1/10th of a second after it is presented in the scene, lining up with primate reaction studies mentioned earlier. (Li et al., 2009) (Yamins and DiCarlo, 2011). These conclusions are the results of years of neuroscience research, and modern understanding of IT neurons indicates they are responsible for the detection of specific complex objects, but rather respond to a variety of visual information. (Desimone et al., 1984) Taken together, it seems that the IT neuron population maintains an explicit representation of the specific identities of objects in a scene, and the question of how this achieved computationally is still not well understood, though there has been work to constrain and determine these algorithmic paradigms. (DiCarlo et al., 2012) (Hung et al., 2005)

Deep Learning Methods

A significant area of interest to both computer scientists and neuroscientists is the question of what computational paradigms, both in the brain and otherwise, allow for efficient learning. The field of machine learning has had numerous breakthroughs, in both statistical and representation learning methods. Machine learning forms a critical part of many modern technologies, including search, social networks, content streams, computer vision systems, and is also making strides in numerous industry verticals including medicine, law, and even policy. (LeCun et al., 2015) Recent years have seen an emphasis



on representation learning, where unprocessed data is inputted into a machine and specific representations are discovered and used for regression or classification. Deep learning, a subset of these representation learning methods, make use of non linear computation across multiple processing nodes that are connected in a graph like fashion.Neuroscience has played an important role in the design of these deep learning systems, also known as neural networks.(Hassabis et al., 2017) Studies of neural computation drove the first mathematical models of neurons, that were then incorporated into artificial neural networks that proved to be useful in modeling a variety of logical functions. It should be noted that nodes in a neural network are gross simplifications of the diversity and sophistication of biological neurons. (Douglas Martin, 1991). Modern deep learning algorithms use multiple representation levels that are transformed by layers of these neurons, very loosely following the hierarchical feed forward patterns of the visual pathway in the brain. Deep networks have shown numerous functions of varying complexity can be learned with enough of these representation transformations. (LeCun et al., 2015) (Hassabis et al., 2017)

An easy to understand example of this is present in image classification domains, where pixels are mapped onto the first layer of artificial neurons in a neural network, where the representation really indicates whether or not there are pixels in a region. The second layer forms a representation of collections or groupings of pixels in a particular format, and later layers form edges, shapes, and eventually whole representations of objects in the images or the whole image itself. These representations can be hand designed, but the true power of learning algorithms is the ability to automate this



representation learning using a specific learning procedure, which is expanded on later. (LeCun et al., 2015) (Ba, Mnih, & Kavukcuoglu, 2014) Many modern applications of these deep neural networks lie in the domain of supervised learning, where the model is trained on data that has an objective label. (Goodfellow et al., 2016) The model is shown the data and computes an output, and is tuned by using an objective function to measure the error difference between the output the model returns and the true score. Learning is accomplished by modifying various internal model parameters to reduce the error. In a neural network, these parameters form the connections between neurons, called weights. (Goodfellow et al., 2016) Understood differently, they are the value of the edges of graph, connecting each node in the network. Sufficiently deep networks often have thousands to millions of these weights, and are trained with similarly large datasets. The adjustment of these weights is accomplished through operations using the gradient vector of the objective. The objective forms a landscape in the n-dimensional space, and the gradient informs us in which direction the minima of this landscape is. Using this, a program can computationally increment towards the minima and thus minimize the desired error as computed by the objective function. (LeCun et al., 2015) Automatic training of multilayer artificial neural networks is done through the use of gradient descent, assuming the network is a smooth function of its inputs and weights. Backpropagation is a procedure that allows researchers to automatically compute the gradient of a selected objective function as described earlier. Backpropagation is essentially an application of the chain rule for derivatives, and can be used on each consecutive layer to adjust weights across the network. Numerous deep learning architectures are present in the literature,



and many comprise of unique, mathematically derived computational rules. (Geron, 2017) Convolutional neural networks operate on data inputs that are formed by multiple arrays, with the popular example being images. They make use of unique layer types that perform convolution operations and pooling operations, and often also include feedforward layers. Other network examples include Recurrent Neural Networks, or RNNs, which process inputs in a ordered fashion, and have hidden neuron units in their layers maintaining a sort of state in the network that allows them to hold information about past elements in the sequence. Extensions of these various networks are present in the literature. For example, an LSTM is an extension of an RNN more focused on the state vector operations. (Goodfellow et al., 2016) (Geron, 2017)

Unsupervised Learning Methods

Many of these architectures can also be used in an unsupervised format, where the learning network is presented with unlabeled data, and the learning objective is usually different. Unsupervised network has been important to recent interest in deep learning, and although the focus of many breakthroughs has been supervised learning, unsupervised is equally promising. Human learning is largely unsupervised, there are no million examples with convenient labels being presented to us in an everyday context. (LeCun et al., 2015) (Hinton et al., 1994) Autoencoders are an example of a deep neural network working in an unsupervised context, and they form a fundamental tool in the experiments proposed in this document. Autoencoders are simple networks that attempt



to reconstruct inputs with as little noise as possible, designed originally with the intention of demonstrating backpropagation usefulness in an unsupervised learning situation. (Goodfellow et al., 2016) In an autoencoder, the inputs themselves are used as learning examples, and the outputs are scored based on their similarity to these inputs. The objective function used to train the autoencoder follows this system, and accordingly the network can be trained using gradient descent as described earlier(Geron, 2017). Autoencoders are of primary interest for the research outlined here because they have been essential in addressing how biological synapses are changed and coordinated in a learning setting. Autoencoders have also been useful as upstream components in supervised learning, where they are stacked and used to inform and tune a larger architecture, yielding excellent results in a variety of tasks over the past few years.(Geron, 2017)

Human Performance vs Deep Neural Networks

With our understanding of core object recognition in the brain and of general deep neural network architectures, it is useful to examine explicit comparative research between the two. As mentioned before, in recent years, DNNs have demonstrated human level object identification after being trained with large amounts of labeled data. (Geron et al., 2017) Research into the introduction of adversarial examples and their role in thorough behavioral comparisons has demonstrated that these deep neural networks can be fooled by modifications that do not explicitly appear to humans completing the same



identification task. (Goodfellow et al., 2016) A recent study conducted a behavioural comparison between the two, reliably demonstrating that DNNs are not just vulnerable to adversarial attacks, but also to random perturbations in image inputs. Many popular architectures, including VGG-16 and AlexNet, demonstrate marked decline in performance with image degradation that doesn't affect human observers. (Geirhos et al., 2017)

Disentangled Representations

While the successes of deep learning methods have been numerous, their performance suffers from lack of robustness and generalisability to new tasks, hallmark traits of Biological Intelligence. On top of this, these learning algorithms are usually extremely data inefficient, and often overfit to the task on which they are being trained. Significant work has been done in tackling these problems from various approaches. Data augmentation is a well explored topic aiming to workaround data inefficiency through the creation of additional synthetic data examples. Recent advances in Deep Learning architectures mentioned earlier operate on the premise that introducing certain biases into the architecture is useful in helping the architecture learn inherent structure in the data. Various forms of inductive biases have been proposed and incorporated in model architectures, with the introduction of convolutional layers and recurrent connections in vision and natural language domains, and more recent breakthroughs like CapsuleNets (Sabour et al., 2017) and Graph Neural Networks (LeCun et al., 2015).



Another alternative is to learn features that generalize to a variety of tasks. There has been a particular focus on learning explicit representations that are faithful to data generative factors, known commonly as learning disentangled representations. While a common definition of disentangled representations, various perspectives have been proposed, including the high level categorical identification perspective introduced earlier in this thesis. These include the idea that single latent encoding units correspond to individual and independent generative factors, or that generative factors may be represented with multiple dimensions. Disentangled representations have been studied in both a supervised and unsupervised context, but the latter constitutes a more realistic training setting faithful to real world scenarios where labeled data is scarce.

Key to the notion of disentangled representations are the idea that, while the data sampled from the real world appears to be extremely high dimensional, a small number of latent, or hidden, factors are responsible for their variations. This relationship between latent data generative factors and real world data is exploited by both artificial and biological information processing systems to learn representations that map to latent factors (Bengio, Courville, & Vincent, 2013). In theoretical neuroscience, the sparse coding model posits that neurons in the primary visual cortex encode for oriented edges, which are latent variables in natural visual scenes. In machine learning, deep autoencoder architectures are used to extract latent variables in the input distribution. The encoding neurons in the bottlenecks of these architectures impose a data limiting factor and are low dimensional. (Hinton & Zemel, 1994; Alemi et al., 2018).



Given their ubiquity, natural questions follow as to what the desirable utility of latent variables is in modelling tasks. Perhaps the most desirable quality is that latent variables should be able to generate data similar to the observed distribution, mapping to the true data generative factors in the real world. Operating under an ideal framework for understanding the nature of disentangled representations is an open question. The research in this paper examines unsupervised learning of disentangled representations from both a probabilistic perspective, specifically in the context of variational inference and VAE, a generalization of the Variational Autoencoder (VAE) (Kingma & Welling, 2013), β-VAE, developed specifically for disentangled representation learning.

A Probabilistic Perspective on Disentangled Representations

In the research outlined in this thesis, we will adopt a probabilistic perspective on modelling latent variables of data distributions (Kingma & Welling, 2019). Here, we can define a generative model $p_\theta(\mathbf{x}, \mathbf{z})$ We are assuming this is the generative process for the data $\mathbf{x}$ and latent variables $\mathbf{z}$.

$$p_{\boldsymbol{\theta}}(\mathbf{x}, \mathbf{z}) = \ p_{\boldsymbol{\theta}}(\mathbf{x}|\mathbf{z})p(\mathbf{z}). \tag{1}$$

where θ denotes the parameters of the generative model. Here, $p(\mathbf{z})$ represents the prior on the latent variables and $p_\theta(\mathbf{x}, \mathbf{z})$ models the data generative process given those latent variables. As mentioned before, there are numerous perspectives and



definitions of disentangled representations, but a useful and intuitive formalism is that the prior on the latent variables, $p(\mathbf{z})$, is factorized.

$$p(\mathbf{z}) = \prod_{i=1}^{k} p_i(z_i).$$ (2)

Statistical independence is strongly implied by factorized distributions, and there are numerous models that have factorized, independent priors like those being discussed, including Independent Component Analysis (ICA)(Hyvarinen & Oja, 2000; Khemakhem, Kingma, & Hyvarinen, 2019). The extraction of statistically independent data generative factors comprises a natural definition of learning disentangled representations (Bengio et al., 2013) used in this line of research, though it is useful to note that a universal definition has not yet been agreed upon in the research community (Bengio et al., 2013; Kingma & Welling, 2013; Locatello et al., 2018; Higgins et al., 2018). A statistically independent representation would carry only information that is useful to the generation of the data, and would carry no other less relevant information (Dayan, Abbott, et al., 2001).

Technically, while the posterior distribution of the model introduced earlier $p_\theta(\mathbf{z}|\mathbf{x})$ allows for the inference of true latent variables, it is often intractable to calculate the model posterior because of the nature of the the the data $\mathbf{x}$. So, while this could theoretically be used to form a disentangled representation, estimation of the model posterior through variational inference methods proves far more useful and efficient in practice.



Variational Inference and Variational Autoencoders

While the successes of deep learning methods have been numerous, their performance suffers from lack of robustness and generalisability to new tasks, hallmark traits of Biological Intelligence. Inference in probabilistic models is often an intractable calculation, and while there are numerous algorithms that provide approximate solutions using sampling, like Markov Chain Monte-Carlo estimation, they have some issues around finding good solutions in a finite horizon and choosing an optimal sampling technique. Here, we will overview a set of methods from Variational Inference that allow for the optimization over a set of tractable distributions as a substitute.

In the Bayesian framework discussed here, calculating the latent from Bayes' Rule is difficult because the data $\mathbf{x}$ is difficult to ascertain. Variational techniques provide a useful workaround by reframing inference as an optimization problem. Rather than try to solve for an intractable probability distribution $p$, variational methods instead optimize over a set of tractable distributions $Q$ to find one similar to $p$. The most similar distribution from the set of tractable distributions $q \in Q$ is then used in place of $p$. Framed in the context of the bayesian perspective we are considering, the Variational scheme allows us to optimize $q_\phi(\mathbf{z}|\mathbf{x})$ that best approximates our model $p_\theta(\mathbf{z}|\mathbf{x})$. Throughout this text, we will refer to the tractable distribution $q_\phi(\mathbf{z}|\mathbf{x})$ as the inference model.



To reformulate the inference task as an optimization problem, an objective for optimization is required. We use the Kullback-Leibler (KL) Divergence, which captures the similarity between $q_\phi(\mathbf{z}|\mathbf{x})$ and $p_\theta(\mathbf{z}|\mathbf{x})$. The KL divergence between two distributions, $q$ and $p$, can be defined as

$$D_{KL}(q||p) = \sum_x q(x) log \frac{q(x)}{p(x)} \tag{3}$$

The KL Divergence constitutes the inference error for the model, $q_\phi(\mathbf{z}|\mathbf{x})$. Throughout this paper, we will be discussing this error term, which we refer to as MIE, as well as the True Inference Error of the generative process, which we will refer to as TIE.

$$\text{MIE} \equiv \mathbb{E}_{p(\mathbf{x})} \left[ D_{KL}(q_\phi(\mathbf{z}|\mathbf{x})||p_\theta(\mathbf{z}|\mathbf{x})) \right]. \tag{4}$$

$$\text{TIE} \equiv \mathbb{E}_{p(\mathbf{x})} \left[ D_{KL}(q_\phi(\mathbf{z}|\mathbf{x})||p_{\text{g-t}}(\mathbf{z}|\mathbf{x})) \right], \tag{5}$$

This distinction is important, as the TIE can only be known if both the ground-truth posterior, which we refer to as $p_{\text{g-t}}(\mathbf{z}|\mathbf{x})$ and the true data generative process from the prior $\mathbf{z}$ are known.

Variational Autoencoders are a class of probabilistic models that makes use of neural networks to learn latent representations. They've achieved success in a wide variety of generative tasks, from vision to language generation. VAEs simultaneously fit



the variational distribution and the parameters of the probabilistic model through the following identity.

$$\ln p_{\boldsymbol{\theta}}(\mathbf{x}) - D_{KL}(q_{\boldsymbol{\phi}}(\mathbf{z}|\mathbf{x})||p_{\boldsymbol{\theta}}(\mathbf{z}|\mathbf{x}))$$
$$= \mathbb{E}_{q_{\boldsymbol{\phi}}(\mathbf{z}|\mathbf{x})}\left[\ln p_{\boldsymbol{\theta}}(\mathbf{x}|\mathbf{z})\right] - D_{KL}(q_{\boldsymbol{\phi}}(\mathbf{z}|\mathbf{x})||p(\mathbf{z})) \tag{6}$$

$\ln p_{\theta}(\mathbf{x})$ defines the log likelihood of the data $\mathbf{x}$ under the parameters of the probabilistic model. The KL divergence, $D_{KL}(q_{\phi}(\mathbf{z}|\mathbf{x})||\mathbf{p}_{\boldsymbol{\theta}}(\mathbf{z}|\mathbf{x}))$ is non negative, which means the right hand side of the equation can serve as a lower bound for the log likelihood of the data, $\ln p_{\theta}(\mathbf{x})$. This relationship can be understood in the following expression:

$$\log p(\mathbf{x}) \geq \mathbb{E}_{q_{\boldsymbol{\phi}}(\mathbf{z}|\mathbf{x})}\left[\log p_{\boldsymbol{\theta}}(\mathbf{x}|\mathbf{z})\right] - D_{KL}\left(q_{\boldsymbol{\phi}}(\mathbf{z}|\mathbf{x})||p(\mathbf{z})\right). \tag{7}$$

This term is known as the Evidence Lower Bound (ELBO).

$$\mathrm{ELBO}(\boldsymbol{\theta}, \boldsymbol{\phi}) \equiv \mathbb{E}_{q_{\boldsymbol{\phi}}(\mathbf{z}|\mathbf{x})}\left[\log p_{\boldsymbol{\theta}}(\mathbf{x}|\mathbf{z})\right] - D_{KL}\left(q_{\boldsymbol{\phi}}(\mathbf{z}|\mathbf{x})||p(\mathbf{z})\right) \tag{8}$$



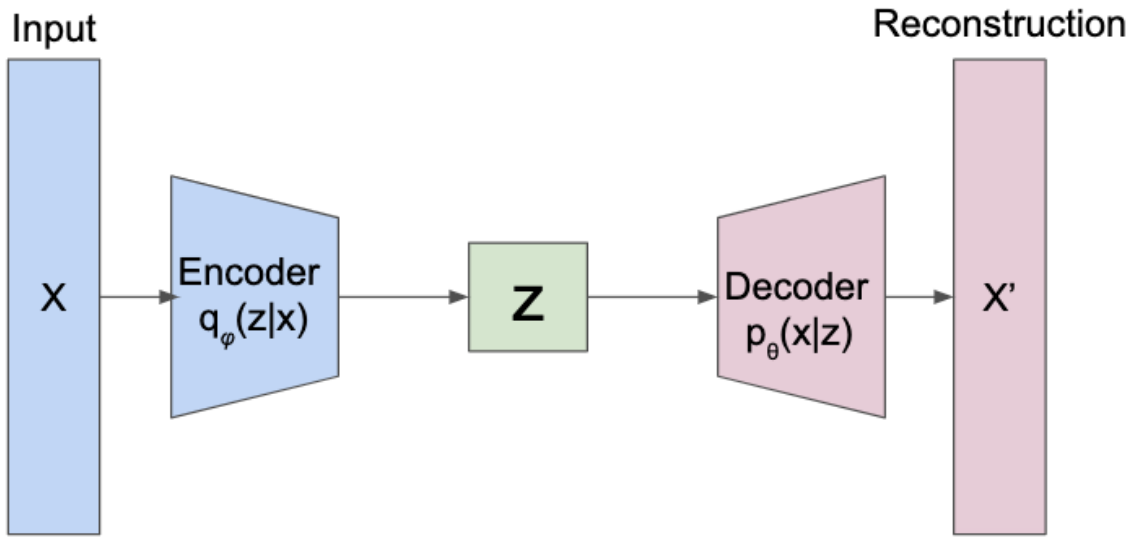

Figure 1: Basic Variational Autoencoder structure. The Encoder and Decoder portions of the network, usually parameterized by neural networks, are depicted as blue and pink respectively. The Latent space encoded by the bottleneck is represented in green.

Variational Autoencoders use neural networks to realize both distributions, $q_\phi(\mathbf{z}|\mathbf{x})$ and $p_\theta(\mathbf{x}|\mathbf{z})$ and maximizes ELBO in the place of data likelihood. The neural network parameterizing the inference model $q_\phi(\mathbf{z}|\mathbf{x})$ is known as the encoder, and the network parameterizing $p_\theta(\mathbf{x}|\mathbf{z})$ is known as a decoder. Similar to the discrete autoencoder architectures discussed earlier, the outputs of the encoder form a low information bottleneck comprised of few neurons. These bottleneck neurons represent latent variables that are inferred by the model. The decoder is used to reconstruct or generate new samples from the VAE's modelled data distribution. Reconstruction accuracy is measured through the left term in ELBO, $\mathbb{E}_{q_\phi(\mathbf{z}|\mathbf{x})}\left[\log p_\theta(\mathbf{x}|\mathbf{z})\right]$. Gradient Descent can be used to optimize the parameters of these two symmetrical networks.



β-VAE is the current state of the art in disentangling latent representations, and an extension to the traditional VAE architecture. In β-VAE, an extra, adjustable hyperparameter β is placed in the objective as a multiplier of the KL Divergence of the inference model and the prior.

$$\mathcal{L}(\boldsymbol{\theta}, \boldsymbol{\phi}; \beta) = \mathbb{E}_{q_{\boldsymbol{\phi}}(\mathbf{z}|\mathbf{x})} \left[ \log p_{\boldsymbol{\theta}}(\mathbf{x}|\mathbf{z}) \right] - \beta D_{KL} \left( q_{\boldsymbol{\phi}}(\mathbf{z}|\mathbf{x}) \| p(\mathbf{z}) \right) \tag{9}$$

## Examining Disentangling in β-VAE

The results outlined in this thesis are the result of a 9 month research project exploring the definition, theoretical implications, and real world simulations of disentangling representations in artificial neural network based learning systems. The goal of this work was to better understand what is meant by disentangled representations, and examine current systems to see if and how they can better achieve this. As mentioned before, the focus is on variational inference in the context of β-VAE, a state of the art inference model for learning disentangled representations. β-VAE includes a modified training objective that enforces conditional independence of learned representations in the encoder's bottleneck. This relies on a strong assumption that the ground source data generating variables are conditionally independent.

A far more natural assumption for latent variables is that they are fully statistically independent. This is important, since statistically independent latent factors



are in general not conditionally independent. The work outlined here is important as it focuses on understanding a very popular research direction in the goal of learning statistically independent disentangled latent variables. The first contribution of this work is general results about the nature of variational inference in the $\beta$-VAE architecture. We demonstrate that increasing $\beta$ leads to a non increasing objective for $\beta$-VAE. This leads to better conditionally independent representations in the encoder bottleneck, but worse reconstruction performance. We also put forth that latent variable inference performance generally tends to be non-monotonic in $\beta$.

The second contribution of this work is the introduction of an analytically tractable $\beta$-VAE model. With this model, we calculate the conditions for an optimal $\beta$ and find that there is an optimal $\beta$ value for inference of latent variables. The model is specialized to the definition of disentangled representations, namely the statistical independence of generative factors.

Finally, we employ deep neural network simulations on multidimensional synthetic visual datasets to test our insights from both the general theorems and our analytical model. We find that simulations agree without theoretical insights. The setup for each of these aims, both theoretical and experimental, is explained in detail in the Methods portion of this text, which immediately follows. The outcome of this research project is explored in the subsequent Results portion of this text. As this research is mathematically involved, a table of frequently used terms is included in Appendix 6. Various derivations and proofs are also included in Appendices 1 through 5.



Chapter II.

Methods

The research aims outlined here were designed to examine the nature of disentangled representations in the β-VAE architecture. There were numerous research methods that contributed to this overarching goal, involving both theoretical and experimental approaches. Simulations and corresponding models are developed using the Python programming language and other ecosystem tools, explained in detail in the tooling section that follows. Data generation paradigms, pre and post processing methods, and limitations are also discussed in detail.

## Methods for General Statements about βVAE Theory

The first research aim is to theoretically understand the effect of the β hyperparameter on the learning of representations and generation of accurate data samples in the β-VAE architecture. The methods employed here were analytical and chiefly involved examining the β-VAE identity and loss functions, introduced in the Variational Inference tutorial, and applying various techniques from the fields of Linear Algebra, Probability, Statistics, and Information Theory. Analytical statements about the nature of β and its dynamics



support and extend some patterns that have been observed in simulation, and are of significant value when establishing the theoretical basis of a research inquiry.

The analysis is primarily done on the identity mentioned above:

$$\ln p_{\boldsymbol{\theta}}(\mathbf{x}) - D_{KL}(q_{\boldsymbol{\phi}}(\mathbf{z}|\mathbf{x})\|p_{\boldsymbol{\theta}}(\mathbf{z}|\mathbf{x}))$$
$$= \mathbb{E}_{q_{\boldsymbol{\phi}}(\mathbf{z}|\mathbf{x})}\left[\ln p_{\boldsymbol{\theta}}(\mathbf{x}|\mathbf{z})\right] - D_{KL}(q_{\boldsymbol{\phi}}(\mathbf{z}|\mathbf{x})\|p(\mathbf{z})) \tag{10}$$

The term is examined through the lens of optimization with regards to $\theta^*$ and $\phi^*$ for the decoder $p_{\theta}(\mathbf{x}|\mathbf{z})$ and encoder $q_{\phi}(\mathbf{z}|\mathbf{x})$, respectively. These terms are given as a solution to

$$\frac{\partial \mathcal{L}}{\partial \boldsymbol{\theta}} = \mathbf{0}, \qquad \frac{\partial \mathcal{L}}{\partial \boldsymbol{\phi}} = \mathbf{0}. \tag{11}$$

The propositions and general statements that resulted from this analysis of β dynamics are presented in detail in the Results section.

## Constructing an Analytically Tractable β-VAE

The next research aim is to demonstrate the general theory in multiple analytically solvable cases. The methods used in this are largely the same set of analytical tools, including principles from linear algebra, probability theory, and calculus, that were used in the general statements provided earlier. Here, the task involves setting up an



analytically tractable version of the β-VAE to better understand the dynamics of the β

hyperparameter. The ground truth posterior is derived for a specific data generative

process, outlined in the results section.

The encoder in this setup contains a fully connected linear neural layer that codes for the

mean of the latent variables, as well as a fully connected linear layer with an exponential

activation function that encodes for the diagonal of the covariance matrix of the derived

posterior. The decoder also consists of a single fully connected linear layer. The

analytically tractable model is also solved numerically to show dynamics of β on a

controlled data generative process involving a simple reconstruction task. The process

used to solve this involved solving numerically solving the derived set of equations

mentioned earlier for a specific autoencoder setup, the setup and using the optimal

parameters from the network to calculate the error terms. This entire process is described

in great detail in the results portion of this text.

Realistic Simulation Methods

Model Architecture

The deep neural network models that were used to parameterize the encoder and decoder

in the numerical experiments used a fixed architecture that was the result of exploration



with various hyperparameters and architecture topologies. The encoder consists of a feed forward network with 3 hidden layers, composed of 256, 200, and 200 units respectively. Similar to the general structure in the analytically tractable model presented earlier, 2 parallel hidden layers with 2 neurons parameterizing the mean and variance for k = 2 sources. The decoder was a symmetrical neural network, and also consisted of 3 feed-forward hidden layers with 200, 200, and 256 units. It is responsible for outputting the reconstructed image.

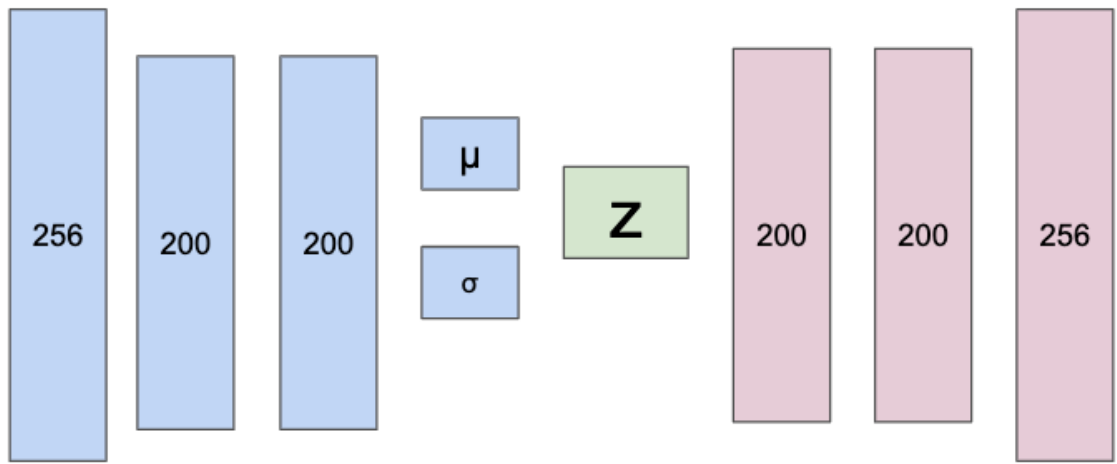

Figure 2: High level architecture used in simulation experiments. Multiple feedforward layers culminate in an encoding representation using the mean and variance of a latent distribution. This distribution is sampled and fed through the symmetrical decoder to generate outputs.



## Training Details

The model was trained for 1000 epochs over the entire 1000 example dataset, and used a tanh activation for the nonlinearities. Optimization was done using using the Adam Optimization algorithm (Kingma & Ba, 2014) with a learning rate of 1e-3. Experiments were repeated across 300 realizations for each β value, and results shown in subsequent sections were averaged over the whole set of realizations.

## Calculation of Objective Terms

To calculate the Reconstruction Objective, 1000 samples were generated from the encoder and passed to the decoder to calculate $\mathbb{E}_{q_{phi}(\mathbf{z}|\mathbf{x})}\left[\log p_{theta}(\mathbf{x}|\mathbf{z})\right]$, averaged over the data $\mathbf{x}$. The Conditional Dependence Loss was calculated directly using the Tensorflow Distributions library's native KL Divergence method. ELBO was calculated by taking the difference of these two terms. Similarly, the β-VAE objective was calculated as the difference with the β value incorporated in the term.

The modelled $\mu_{\mathbf{z}}$ and $\sigma_{\mathbf{z}}$ were used to numerically calculate the Inference Error, using an estimation of the data distribution from the batches.



Data Processing and Generation

The deep β-VAE simulations made use of multiple synthetic datasets. These datasets involved pasting an MNIST digit onto a blank 40 x 40 pixel canvas where pixel values were 0. Multiple datasets and experiments were composed, some involving localizing a single MNIST digit onto the canvas based on a data generative scheme involving independent sources, in this case the x and y coordinates corresponding to vertical and horizontal locations, being mixed and used as the position for the digit on the canvas. These results are briefly explored and included in the Results section. The final experimental setup involved mixing multiple mnist digits using a similar generative scheme discussed in the results. Digits were mixed and localized on a blank canvas, and the reconstruction scheme was used. Resulting images could be interpreted as interpolations of the two mixing digits used.

The dataset required minimal preprocessing other than the standardization that centers the data at mean 0 with standard deviation 1. This is because MNIST is already optimized for usage in neural networks with each image 28 x 28 pixels in size. Because of the nature of these small images, no convolution or image sampling has to be done, and the entire image batch can be directly passed to the network. Post training the process is largely the same, with the data moved back to its original distribution to better observe results, loss values calculated, and most importantly the latent space plotted and traversed for each



digit to see if our network is properly capturing separate factors at each bottleneck. Reconstructed images, both those reconstructed from manipulations of the bottleneck neurons and those directly reconstructed from input data, were plotted using the tools described in the following section.

Tools and Resources

Several libraries and toolsets built by the open source and scientific communities are used to build the experimental network architectures outlined in this project. (Oliphant, 2007) The Anaconda Distribution of Python 3 is used, providing a useful package manager, environment manager, and access to several thousand open source scientific computing packages, including all the tools used in this research. Python's operating system library will was used to facilitate various data pipelines, transferring generated data into the scope of the model files to effectively allow for training and inference. Numpy was used as a base data format for vector and matrix manipulations and broadcasting, forming the main data primitive used in the experiments. The library also allows for fast model performance, as it compiles down to lower level languages like C and Fortran. (Van Der Walt et al., 2011) Scipy optimization packages were used for numerical experiments corresponding to the analytically tractable models. (Jones, 2014) The scikit learn machine learning library was used for its preprocessing functions, including scaling and



normalizing data inputs as necessary for optimal neural network performance. Image

creation and manipulation, primarily in the dataset generation module, was achieved

through the use of the OpenCV python interface along with the popular image editing

library PIL. (Oliphant, 2007) The deep networks themselves were designed primarily in

Tensorflow, because of its robustness and high level specification. Tensorflow is a

standard in both experimental and production use cases, and achieves numerical

computation using graphs for data flow. (Geron, 2017) Nodes on the graph are used as

mathematical operations, while weights in the network are represented by

multidimensional tensors making up the edges of the graph. Tensorflow allows for the

use of both CPU and GPU hardware, as well as other IPUs and accelerators, and as such

is an optimal framework choice to design the aforementioned deep architectures. (Geron,

2017) Finally the popular 2D plotting library, Matplotlib, was used to generate figures in

the data analyses provided, specifically for insight into learning, error rates, objective

term behaviors, and the latent space of the autoencoder bottleneck. (Van Der Walt et al.,

2011)



Chapter III.

Results

As the research undertaken in this project had both strong theoretical and empirical components, the results have been broken down according to each research claim. In the first section, general theorems about the nature of variational inference in the context of the β-VAE architecture. Following this we examine the results of the analytical model introduced earlier and use it to inform our understanding of β-VAE behavior with statistically independent latent variables. Following this, the analytical model is calculated numerically. Finally, the results of the realistic deep β-VAE on the synthetic dataset are examined.

Theoretical Statements about β Dynamics in β-VAE

This section comprises the result of theoretically examining the dynamics of the β parameter on the learning of representations in β-VAE. General theoretical statements are presented as the result of proving the effect the hyperparameter β has on various terms in the following identity presented in the tutorial on Variational Inference:

$$
\begin{aligned}
\ln p_{\boldsymbol{\theta}}(\mathbf{x}) &- D_{KL}(q_{\boldsymbol{\phi}}(\mathbf{z}|\mathbf{x})||p_{\boldsymbol{\theta}}(\mathbf{z}|\mathbf{x})) \\
&= \mathbb{E}_{q_{\boldsymbol{\phi}}(\mathbf{z}|\mathbf{x})}\left[\ln p_{\boldsymbol{\theta}}(\mathbf{x}|\mathbf{z})\right] - D_{KL}(q_{\boldsymbol{\phi}}(\mathbf{z}|\mathbf{x})||p(\mathbf{z}))
\end{aligned} \tag{12}
$$



The optimal β-VAE parameters that maximize the objective:

$$\mathcal{L}_{\beta} = \mathbb{E}_{q_{\phi}(\mathbf{z}|\mathbf{x})}\left[\log p_{\boldsymbol{\theta}}(\mathbf{x}|\mathbf{z})\right] - \beta D_{KL}\left(q_{\phi}(\mathbf{z}|\mathbf{x})\|p(\mathbf{z})\right). \tag{13}$$

are denoted $\theta^*$ and $\phi^*$ decoder and encoder, respectively. Framed formally, they are the solution to the following statements:

$$\frac{\partial \mathcal{L}}{\partial \boldsymbol{\theta}} = \mathbf{0}, \qquad \frac{\partial \mathcal{L}}{\partial \boldsymbol{\phi}} = \mathbf{0}. \tag{14}$$

There are several theoretical propositions introduced in this section. The first two are analytical insights that lend understanding into the behavior of β in relation to reconstruction quality. The third gives insight into the Model Inference Error using ELBO. Note that while all the statements are presented with respect to a single data point, they generalize to finite training sets and averages over $p(\mathbf{x})$.

The first proposition involves the behavior of the optimal objective, defined below, as a function of β.

$$\mathcal{L}^*(\beta) \equiv \mathcal{L}(\boldsymbol{\theta}^*(\beta), \boldsymbol{\phi}^*(\beta), \beta), \tag{15}$$

Namely, the optimal value of the β-VAE objective, $\mathcal{L}^*(\beta)$, is non-increasing with increasing β.

$$\frac{\partial \mathcal{L}^*(\beta)}{\partial \beta} = -D_{KL}\left(q_{\boldsymbol{\phi}^*}(\mathbf{z}|\mathbf{x})\|p(\mathbf{z})\right) \leq 0. \tag{16}$$



Based on the conditions for optimality provided above and making use of the chain rule, the following is found:

$$\frac{\partial \mathcal{L}^*}{\partial \beta} = \left( \frac{\partial \mathcal{L}}{\partial \boldsymbol{\theta}} \cdot \frac{\partial \boldsymbol{\theta}}{\partial \beta} + \frac{\partial \mathcal{L}}{\partial \boldsymbol{\phi}} \cdot \frac{\partial \boldsymbol{\phi}}{\partial \beta} + \frac{\partial \mathcal{L}}{\partial \beta} \right) \Bigg|_{\boldsymbol{\theta}=\boldsymbol{\theta}^*, \boldsymbol{\phi}=\boldsymbol{\phi}^*}$$
$$= -D_{KL}(q_{\boldsymbol{\phi}^*}(\mathbf{z}|\mathbf{x}) || p(\mathbf{z})) \leq 0. \tag{17}$$

The second proposition demonstrates the change in the optimal objective change with β. The relevant proof is included in the appendix. With increasing β, the KL divergence between $q_{\phi^*}(\mathbf{z}|\mathbf{x})$ and $p(\mathbf{z})$ is non-increasing.

$$\frac{d}{d\beta} D_{KL}(q_{\phi^*}(\mathbf{z}|\mathbf{x}) || p(\mathbf{z})) \leq 0. \tag{18}$$

Along with the previous insight regarding the non-increasing β-VAE objective,

$$\frac{d \, \mathbb{E}_{q_{\phi^*}(\mathbf{z}|\mathbf{x})} \left[ \log p_{\boldsymbol{\theta}^*}(\mathbf{x}|\mathbf{z}) \right]}{d\beta} \leq 0 \tag{19}$$

The next proposition is regarding the behavior of ELBO at the optimal point with respect to β. The value of ELBO at the optimal point is denoted by

$$\text{ELBO}^*(\beta) \equiv \text{ELBO}(\boldsymbol{\theta}^*(\beta), \boldsymbol{\phi}^*(\beta)). \tag{20}$$



By definition,

$$\mathcal{L} = \text{ELBO} + (1 - \beta) D_{KL}(q_\phi(\mathbf{z}|\mathbf{x}) || p(\mathbf{z})). \qquad (21)$$

From this result and the behavior of the KL divergence between the inference model and the prior with respect to β, $\text{ELBO}^*$ is maximized at β = 1. β affects latent variable inference as well. Recall that this is measured by MIE. Rearranging and evaluating the identity at optimal parameters for β-VAE demonstrates interesting behavior about the β to infinity limit, where the inference model becomes increasingly conditionally independent.

$$\text{MIE}(\beta) = \mathbb{E}_{p(\mathbf{x})} \left[ \ln p_{\boldsymbol{\theta}^*(\beta)}(\mathbf{x}) - \text{ELBO}^*(\beta) \right]. \qquad (23)$$

It can be expected that as reconstruction performance worsens with β, the data likelihood decreases. If the data log-likelihood was monotonic with β, because of the non-monotonic behavior of ELBO with a maximum, MIE can be expected to exhibit non-monotonic behavior with an optimal value. In the next section we see specific examples of this relationship in the context of analytically tractable models.

## Analytically Tractable β-VAE Results

To examine the insights from the theoretical propositions proved regarding the dynamics and effects to the hyperparameter β, an analytically tractable β-VAE was presented. 2 different cases were examined.



The first finding was that fixing the decoder, which parameterized $p_\theta(\mathbf{x}|\mathbf{z})$, does not lead to β-VAE having better disentangling. This is a fairly simple case, and can be intuitively understood as the decoder network not being trained. This would mean fixing our parameters for the decoder, θ. In this case, training the β-VAE objective would only train the encoder network that parametrizes the inference model $q_\phi(\mathbf{z}|\mathbf{x})$. We previously set up the MIE as a function of the hyperparameter β:

$$\text{MIE}(\beta) = \mathbb{E}_{p(\mathbf{x})} \left[ \ln p_{\boldsymbol{\theta}^*(\beta)}(\mathbf{x}) - \text{ELBO}^*(\beta) \right]. \tag{24}$$

on the behavior of ELBO put forth by the third proposition in the previous section, we see that ELBO being maximized at β = 1 would extend to cases with fixed decoder parameter θ. Since ELBO constitutes the lower bound for the MIE, it is clear that MIE is minimum at the same value for β. The data likelihood, represented by $p_\theta(\mathbf{x})$ in the MIE expression, doesn't change as a result of the training process. All together, this would mean that in the case that the decoder in β-VAE is best at learning the true latent variables at a β value of 1. This is the same as the original VAE architecture, and would imply that the extension to β does not improve the learning it aims to. The second set of results focuses on the optimal β values in a general analytically tractable version of β-VAE. Accordingly, each term can be calculated with relation to β in our identity:

$$\begin{aligned}
\ln p_{\boldsymbol{\theta}}(\mathbf{x}) &- D_{KL}(q_{\boldsymbol{\phi}}(\mathbf{z}|\mathbf{x})||p_{\boldsymbol{\theta}}(\mathbf{z}|\mathbf{x})) \\
&= \mathbb{E}_{q_{\boldsymbol{\phi}}(\mathbf{z}|\mathbf{x})} \left[ \ln p_{\boldsymbol{\theta}}(\mathbf{x}|\mathbf{z}) \right] - D_{KL}(q_{\boldsymbol{\phi}}(\mathbf{z}|\mathbf{x})||p(\mathbf{z}))
\end{aligned} \tag{25}$$



For this analytically tractable model, we have certain assumptions around how the data distribution is created. Our data, $\mathbf{x}$, is generated through the following scheme, detailed earlier in the methods portion of this thesis.

$$\mathbf{x} = \mathbf{A}\mathbf{s} + \boldsymbol{\eta}. \tag{26}$$

Recall that the ground truth factors, $\mathbf{s} \in \mathbb{R}^{\mathbf{k}}$, are mixed through the mixing matrix, $\mathbf{A} \in \mathbb{R}^{\mathbf{N} \times \mathbf{k}}$. $\eta \in \mathbb{R}^{N}$, representing noise, is added to corrupt the mixed result. Here, we assume that $\mathbf{s} \sim (\mathbf{0}, \mathbf{I_k})$ and $\eta \sim \mathbf{N}(\mathbf{0}, \mathbf{I_N})$. The data distribution $p_{\theta}(\mathbf{x})$ is found to be

$$p(\mathbf{x}) = \mathcal{N}(\mathbf{0}, \mathbf{A}\mathbf{A}^{\top} + \mathbf{I}_N) \equiv \mathcal{N}(\mathbf{0}, \boldsymbol{\Sigma_x}). \tag{27}$$

The ground truth posterior can be derived and can be calculated exactly in this model. This derivation is included in the appendix of this text.

$$p_{\text{g-t}}(\mathbf{s}|\mathbf{x}) = \mathcal{N}(\boldsymbol{\mu_{s|x}}, \boldsymbol{\Sigma_{s|x}}),$$
$$\text{with } \boldsymbol{\mu_{s|x}} = (\mathbf{A}^{\top}\mathbf{A} + \mathbf{I}_k)^{-1}\mathbf{A}^{\top}\mathbf{x}$$
$$\text{and } \boldsymbol{\Sigma_{s|x}} = (\mathbf{A}^{\top}\mathbf{A} + \mathbf{I}_k)^{-1}. \tag{28}$$

Note that $I_d$ represents a d × d identity matrix. The latent factors in this model, while statistically independent, are conditionally dependent when conditioned on data, which is the case here. Thus, the covariance matrix in the posterior is non-diagonal. Accordingly, the MIE and TIE are dependent on β.



Recall from the methods section that in this analytical approach, the encoder that parametrizes the inference model, $q_\phi(\mathbf{z}|\mathbf{x})$, includes a fully connected layer. This layer, $\{\mathbf{W}^\mu, \mathbf{b}^\mu\}$, encodes the mean, $\mu_\mathbf{z}$, and is linearly activated. Another similarly fully connected layer, $\{\mathbf{W}^\sigma, \mathbf{b}^\sigma\}$, encodes the diagonal of the covariance matrix $\Sigma_z$. This layer is exponentially activated. Given an input $\mathbf{x}$, the network provided can generate the mean of the latent variables:

$$\boldsymbol{\mu}_\mathbf{z} = \mathbf{W}^\mu \mathbf{x} + \mathbf{b}^\mu \tag{29}$$

as well as the variance. Note that the the diag operation maps vectors in $\mathbb{R}^k$ to the diagonal of a $\mathbb{R}^{k \times k}$ diagonals matrix.

$$\boldsymbol{\Sigma}_\mathbf{z} = \mathrm{diag}(\exp(\mathbf{W}^\sigma \mathbf{x} + \mathbf{b}^\sigma)), \tag{30}$$

exponential activation involved in the encoding of the covariance matrix is performed on each element, and results in only nonnegative covariances. The decoder of this model parameterized $p_\theta(\mathbf{x}|\mathbf{z})$, and using a gaussian prior $p(\mathbf{z}) = \mathcal{N}(\mathbf{0}, \mathbf{I_k})$, we can calculate the data likelihood through the following integral:

$$p_{\boldsymbol{\theta}}(\mathbf{x}) = \int d\mathbf{z}\, p_{\boldsymbol{\theta}}(\mathbf{x}|\mathbf{z}) p(\mathbf{z}) \tag{31}$$

The decoder here is a single fully connected layer with linear activation, $\{\mathbf{D}, \mathbf{b}^\mathbf{D}\}$. With the assumption that the output of the decoder is normally distributed,



$$\mathbf{y} \sim \mathcal{N}(\mathbf{Dz} + \mathbf{b}^D, \sigma_y^2 \mathbf{I}_N) \tag{32}$$

where $\sigma_y^2$ is a hyperparameter we fix to be 1.

This is important, as now the data generative process outlined earlier can be fully modelled by the decoder in the case $\mathbf{D} = \mathbf{A}$, $\mathbf{b}^\mathbf{D} = \mathbf{0}$ and $\sigma_y^2 = 1$. Now, any difference from these parameter choices is the fault of the encoder not capturing the ground truth distribution.

Recall the β-VAE objective introduced earlier:

$$\mathcal{L}_\beta = \mathbb{E}_{q_\phi(\mathbf{z}|\mathbf{x})}\left[\log p_{\boldsymbol{\theta}}(\mathbf{x}|\mathbf{z})\right] - \beta D_{KL}\left(q_\phi(\mathbf{z}|\mathbf{x})\|p(\mathbf{z})\right). \tag{33}$$

Using the data distribution defined earlier, the data can be integrated out of the objective. The result of this integration is the expanded loss function below. The appendix includes the full integration.

$$\begin{aligned}
\mathcal{L}_\beta = -\frac{1}{2}\Bigg\{ &\mathrm{Tr}\left[(\mathbf{DW}^\mu - \mathbf{I}_N)\boldsymbol{\Sigma}_\mathbf{x}(\mathbf{DW}^\mu - \mathbf{I}_N)^\top\right] \\
&+ \beta\mathrm{Tr}\left[\mathbf{W}^\mu\boldsymbol{\Sigma}_\mathbf{x}(\mathbf{W}^\mu)^\top\right] \\
&+ \sum_i^k \left(\left[\mathbf{D}^\top\mathbf{D}\right]_{ii} + \beta\right) e^{\frac{1}{2}\left[\mathbf{W}^\sigma\boldsymbol{\Sigma}_\mathbf{x}(\mathbf{W}^\sigma)^\top\right]_{ii} + b_i^\sigma} \\
&+ (\mathbf{Db}^\mu + \mathbf{b}^D)^2 + \beta(\mathbf{b}^\mu)^2 - \beta\sum_i^k b_i^\sigma \Bigg\}. \tag{34}
\end{aligned}$$

Optimizing over the parameters of the network in the analytical case amounts to setting the partial derivative of the loss above with respect to $\{\mathbf{W}^\mu, \mathbf{b}^{\mu,\sigma}, \mathbf{b}^\sigma, \mathbf{D}, \mathbf{b}^\mathbf{D}\}$ to zero. Doing this involves unpacking the indices.



$$L = -\frac{1}{2} \Bigg\{ (D_{ij} W_{jk}^{\mu} - \delta_{ik}) \Sigma_{kl} (W_{ml}^{\mu} D_{im} - \delta_{il}) - \beta \sum_i b_i^{\sigma}$$
$$+ \beta (W^{\mu})_{ij} \Sigma_{jk} W_{ik}^{\mu} + (D_{ij} b_j^{\mu} + b_i^D)^2 + \beta (b_i^{\mu})^2$$
$$+ \sum_i \left( D_{li}^2 + \beta \right) \exp \left( \frac{1}{2} W_{ij}^{\sigma} \Sigma_{jk} W_{ik}^{\sigma} + b_i^{\sigma} \right) \Bigg\} \tag{35}$$

Note that $\Sigma_x$ has been simplified as $\Sigma$. Repeated indices are summed over unless otherwise specified. Following this,

$$0 = \frac{\partial L}{\partial W_{ab}^{\mu}} = \left[ \left( \mathbf{D}^{\top} (\mathbf{D} \mathbf{W}^{\mu} - \mathbf{I}_N) + \beta \mathbf{W}^{\mu} \right) \mathbf{\Sigma} \right]_{ab}, \tag{36}$$

$$0 = \frac{\partial L}{\partial b_a^{\mu}} = \left[ (\mathbf{D} \mathbf{b}^{\mu} + \mathbf{b}^D) \mathbf{D} + \beta \mathbf{b}^{\mu} \right]_a, \tag{37}$$

$$0 = \frac{\partial L}{\partial D_{ab}} = \left[ (\mathbf{D} \mathbf{W}^{\mu} - \mathbf{I}_N) \mathbf{\Sigma} (\mathbf{W}^{\mu})^{\top} \right]_{ab}$$
$$+ \left[ \mathbf{D} \mathbf{b}^{\mu} + \mathbf{b}^D \right]_a b_b^{\mu} + D_{ab} e^{\frac{1}{2} \left[ \mathbf{W}^{\sigma} \mathbf{\Sigma} (\mathbf{W}^{\sigma})^{\top} \right]_{bb} + b_b^{\sigma}}, \tag{38}$$

$$0 = \frac{\partial L}{\partial b_a^D} = \left[ \mathbf{D} \mathbf{b}^{\mu} + \mathbf{b}^D \right]_a \tag{39}$$

$$0 = \frac{\partial L}{\partial W_{ab}^{\sigma}} = \left( \left[ \mathbf{D}^{\top} \mathbf{D} \right]_{aa} + \beta \right) e^{\frac{1}{2} \left[ \mathbf{W}^{\sigma} \mathbf{\Sigma} \right]_{ab} W_{ab}^{\sigma} + b_a^{\sigma}} \left[ \mathbf{W}^{\sigma} \mathbf{\Sigma} \right]_{ab}, \tag{40}$$

$$0 = \frac{\partial L}{\partial b_a^{\sigma}} = \left( \left[ \mathbf{D}^{\top} \mathbf{D} \right]_{aa} + \beta \right) e^{\frac{1}{2} \left[ \mathbf{W}^{\sigma} \mathbf{C} (\mathbf{W}^{\sigma})^{\top} \right]_{aa} + b_a^{\sigma}} - \beta. \tag{41}$$

Examining the equations pertaining to $\mathbf{b}^{\mu}$ and $\mathbf{b}^D$, we see

$$\mathbf{b}^{\mu} = \mathbf{b}^D = 0 \tag{42}$$

The remaining set of equations are ($a = 1, ..., N; \ b = 1, ..., k$):



$$0 = \left[ \left( \mathbf{D}^\top (\mathbf{DW}^\mu - \mathbf{I}_N) + \beta \mathbf{W}^\mu \right) \boldsymbol{\Sigma_x} \right]_{ab},$$
$$0 = \left[ (\mathbf{DW}^\mu - \mathbf{I}_N) \boldsymbol{\Sigma_x} (\mathbf{W}^\mu)^\top \right]_{ab} + D_{ab} e^{\frac{1}{2} \left[ \mathbf{W}^\sigma \boldsymbol{\Sigma_x} (\mathbf{W}^\sigma)^\top \right]_{bb} + b_b^\sigma},$$
$$0 = \left( \left[ \mathbf{D}^\top \mathbf{D} \right]_{aa} + \beta \right) e^{\frac{1}{2} \left[ \mathbf{W}^\sigma \boldsymbol{\Sigma_x} \right]_{ab} W_{ab}^\sigma + b_a^\sigma} \left[ \mathbf{W}^\sigma \boldsymbol{\Sigma_x} \right]_{ab},$$
$$0 = \left( \left[ \mathbf{D}^\top \mathbf{D} \right]_{aa} + \beta \right) e^{\frac{1}{2} \left[ \mathbf{W}^\sigma \boldsymbol{\Sigma_x} (\mathbf{W}^\sigma)^\top \right]_{aa} + b_a^\sigma} - \beta. \tag{43}$$

Using the optimal values for the network, the posterior can be calculated, and we can see that

$$p_\theta(\mathbf{z}|\mathbf{x}) = \mathcal{N}(\boldsymbol{\mu_{z|x}}, \boldsymbol{\Sigma_{z|x}}),$$
$$\text{with } \boldsymbol{\mu_{z|x}} = (\mathbf{D}^\top \mathbf{D} + \mathbf{I}_k)^{-1} \mathbf{D}^\top \mathbf{x}$$
$$\text{and } \boldsymbol{\Sigma_{z|x}} = (\mathbf{D}^\top \mathbf{D} + \mathbf{I}_k)^{-1}. \tag{44}$$

The full derivation of the model posterior is in the appendix. In the case that D = A, the posterior for the model is the same as the true, ground-truth posterior. The focus of this analysis is to understand the behavior of the error terms, both the MIE and TIE. The data can be integrated out for these terms, as demonstrated in the appendix. The inference error is found to be

$$\text{MIE/TIE}$$
$$= \frac{1}{2} \Bigg\{ \sum_i^k E_{ii}^{-1} \exp \left[ \frac{1}{2} \left( \mathbf{W}^\sigma \boldsymbol{\Sigma_x} (\mathbf{W}^\sigma)^\top \right)_{ii} + b_i^\sigma \right] - \sum_i^k b_i^\sigma$$
$$+ \operatorname{Tr} \log \mathbf{E} + \operatorname{Tr} \left[ (\mathbf{F} - \mathbf{W}^\mu)^\top \mathbf{E}^{-1} (\mathbf{F} - \mathbf{W}^\mu) \boldsymbol{\Sigma_x} \right] - k \Bigg\}, \tag{45}$$

For TIE:

$$\mathbf{E} = (\mathbf{A}^\top \mathbf{A} + \mathbf{I}_k)^{-1}, \quad \mathbf{F} = (\mathbf{A}^\top \mathbf{A} + \mathbf{I}_k)^{-1} \mathbf{A}^\top. \tag{46}$$

and for MIE:

$$\mathbf{E} = (\mathbf{D}^\top \mathbf{D} + \mathbf{I}_k)^{-1}, \quad \mathbf{F} = (\mathbf{D}^\top \mathbf{D} + \mathbf{I}_k)^{-1} \mathbf{D}^\top, \tag{47}$$



# Numerical Verification of Analytical Model

Numerical simulations were used to verify insights from the analytical approach outlined earlier. Recall the system of equations presented in (43). In the simulations, this set of equations is solved using $\{\mathbf{W}^{\mu*}, \mathbf{b}^{\mu*}, ^{\sigma*}, \mathbf{b}^{\sigma*}, \mathbf{D}^*, \mathbf{b}^{\mathbf{D}*}\}$ to calculate terms. Recall that these are the optimal network parameters. The terms calculated include the Model Inference Error, the True Inference Error, and ELBO. These terms were explained in the setup for β-VAE in the first portion of this text, and also in the realistic simulation section that follows. The set of equations was solved for $N = 128, k = 2, A_{ij} = 1/2(1 + \delta_{ij})$. The results indicate that while the inference error is not decreasing monotonically and that there is a clear minimum at some β, ELBO is maximized at β = 1. This supports the theoretical results, and agrees with the simulations that follow. The reconstruction objective,

$$\mathbb{E}_{q_{\boldsymbol{\phi}}(\mathbf{z}|\mathbf{x})} \left[ \log p_{\boldsymbol{\theta}}(\mathbf{x}|\mathbf{z}) \right] \tag{49}$$

and the Conditional Independence Loss,

$$D_{KL} \left( q_{\boldsymbol{\phi}}(\mathbf{z}|\mathbf{x}) \| p(\mathbf{z}) \right) \tag{50}$$

was also tracked. These terms also seem to monotonically decrease with the β hyperparameter, agreeing with the theoretical insights.



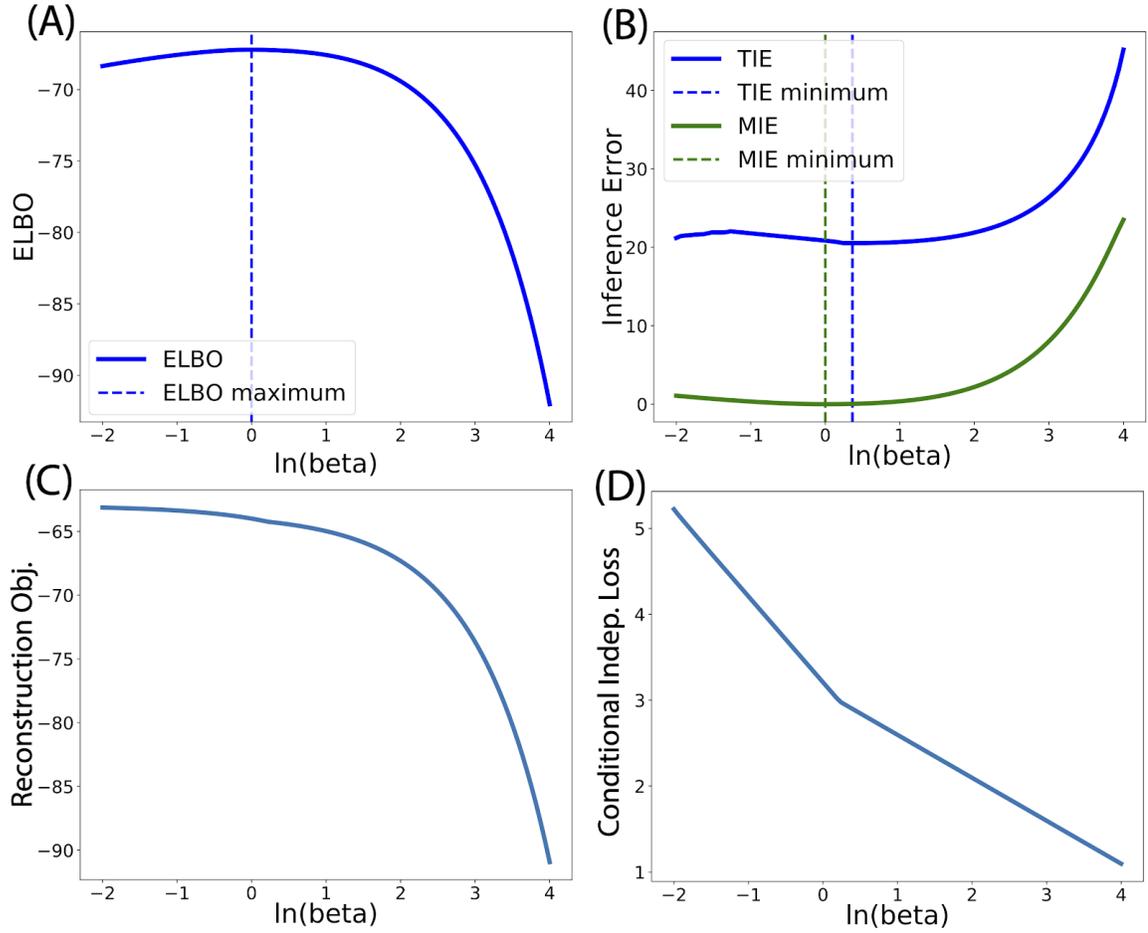

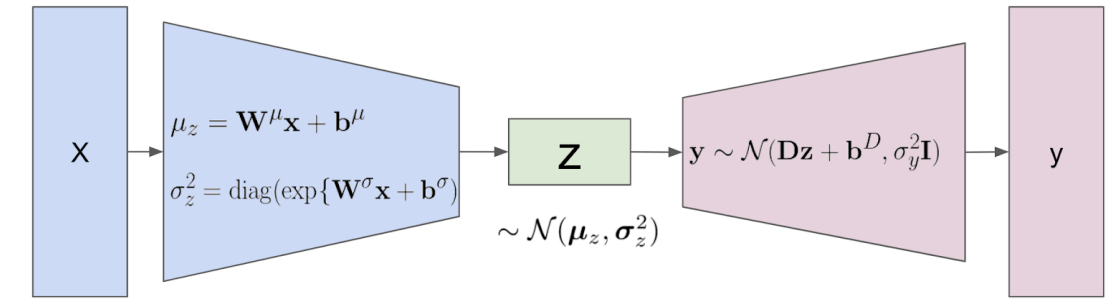

Figure 3: Above: β-dependence of various quantities at the optimal parameter configuration of β-VAE. (A) ELBO as a function of β. Vertical dashed line represent the extremum. (B) MIE/TIE as a function of β. (C) Reconstruction objective as a function of β. (D) Conditional IndependenceLoss as a function of β. In these plots, we averaged the plotted quantities over the data distribution. Below: Diagram of Analytical Model



Realistic Simulation with Deep β-VAE

To examine the theoretical and analytical insights into β dynamics, the deep neural network based simulation outlined in the third aim of the methods portion of this document was executed. Recall that the dataset generation involved the following scheme: x = As + η. The specific experimental setup can be found in the methods section, but the task involved the mixing of 10 different handwritten digits from the MNIST dataset. The encoder and decoder, $q_\phi(\mathbf{z}|\mathbf{x})$ and $p_\theta(\mathbf{x}|\mathbf{z})$, are symmetrical neural networks with three feed forward fully connected layers. Each neuron includes an activation function, in this case tanh. The encoder outputs two separate layers encoding the mean and variance of the latent variables, $\mu_z$ and $\Sigma_z$. With inputs z from the posterior distribution generated by the encoder, the decoder reconstructs realistic data samples.

After training the network with different values of β in the objective, the decoder's reconstructions offer varying results. Specifically, higher β values present in the training objective leads to noticeably worsening reconstruction quality of the digits, though encoding in bottleneck neurons seems to segregate into somewhat factorized representations. Conversely, when trained with lower beta values, reconstruction improves but encoder bottleneck neurons don't contain representations that are distinct. This can be seen in Figure 5.

The simulations also agree with the insights put forth in the theoretical and analytical portions of this thesis regarding the dynamics of the β hyperparameter and its



effects on the various terms in the β-VAE identity. The terms tracked include the Reconstruction Objective defined in the VAE introduction, and the Conditional Independence Loss. Following training, these individual terms were calculated and plotted in Figure 4. The results confirm what was observed in the analytically tractable case and proposed by the theoretical statements, namely that these terms are decreasing with β. The ELBO term, was also tracked and can be observed in the plot. The ELBO maximizes near β = 1, and otherwise decreases with β. The True Inference Error was also calculated for the deep β-VAE architecture, which is possible because the entire experimental setup has been designed, including the data generative process. The TIE follows behaves non-monotonically and an optimal β value exists.

The other experimental setup with a single MNIST digit positioned according to a similarly mixed data generative process also demonstrated interesting results. The behaviors of terms in this simulation were also consistent with the findings presented earlier, and reconstruction quality noticeably worsened as a result of increasing β. Bottleneck neurons in this experimental setup seemed to encode for structured, orthogonal axes of motion following training.



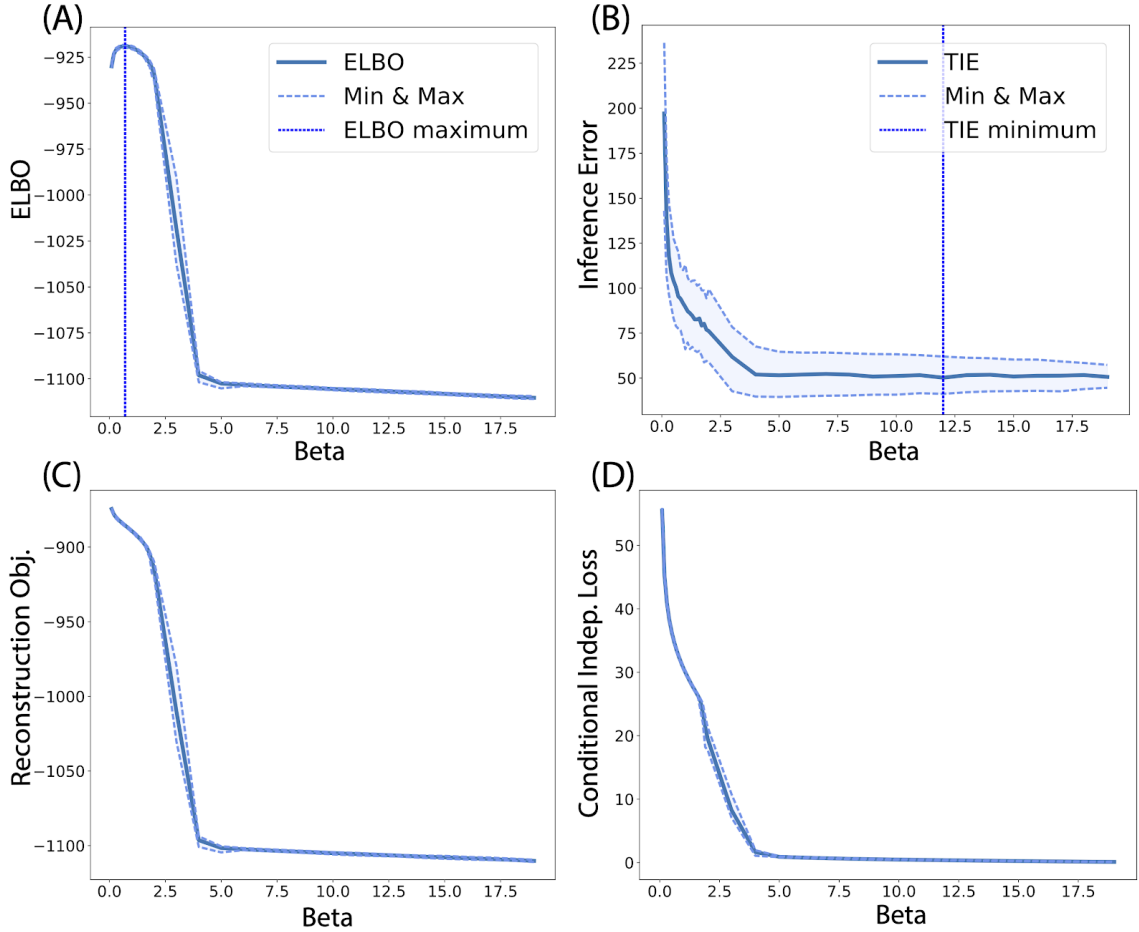

Figure 4: Values for error terms across 100 random initializations of the network. Solid line represents the average. Dashed lines around the solid line represent the minimum and maximum values, and vertical dashed line represent the extremum. (A) ELBO as a function of β. (B) TIE as a function of β. (C) Reconstruction Objective as a function of β. (D) Conditional IndependenceLoss as a function of β



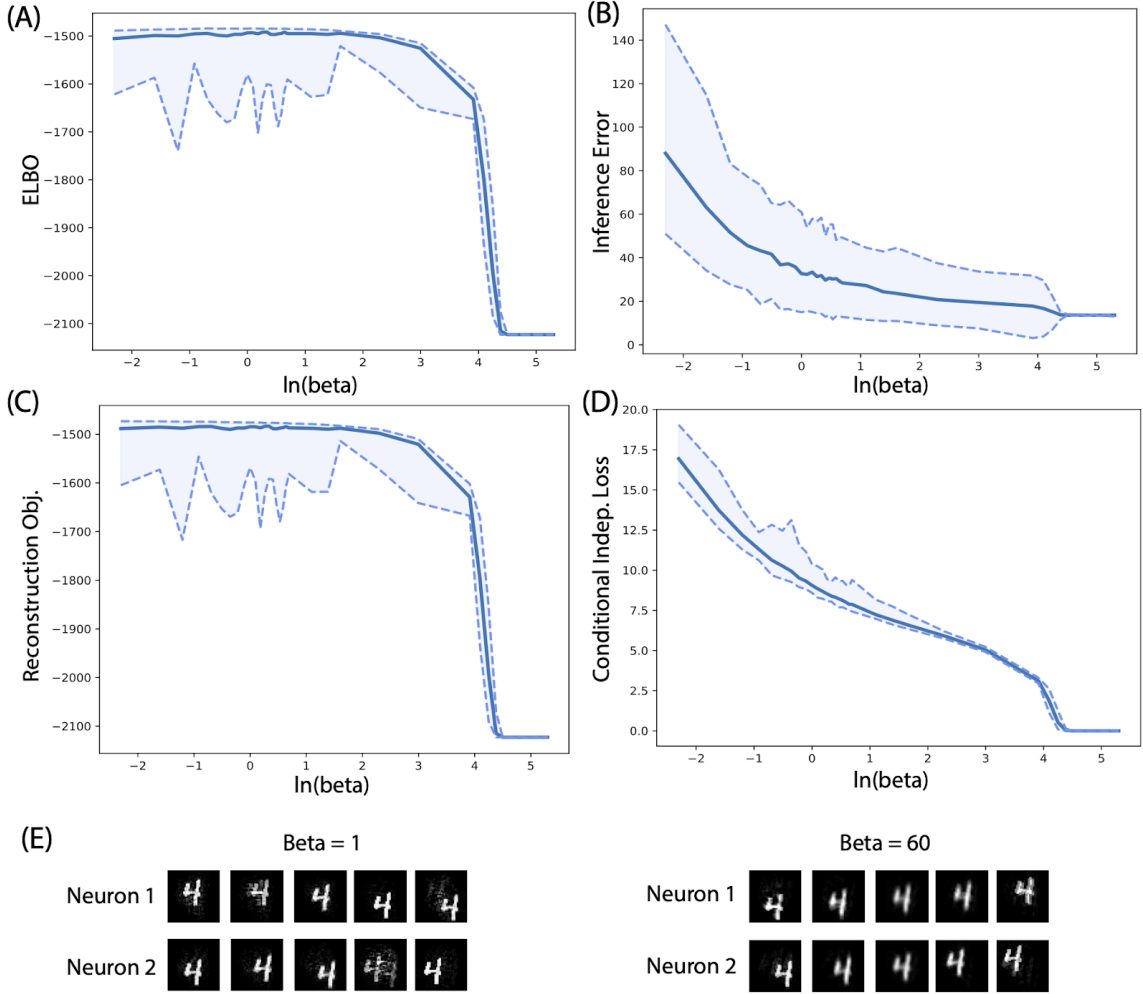

Figure 5: A different synthetic dataset, which comprises of a single MNIST digit localized at different locations on a blank canvas. The cartesian coordinate of the digit in a sample from our data $\mathbf{x}$, is determined by the data generative process, with $A_{ij} = 2\delta_{ij} + 0.73$, $\sim (0,_k)$, $\eta \sim (0,_N)$, $N = k = 2$. Dashed lines represent the minimum and maximum values, and solid line represents the average. (A) ELBO as a function of β. (B) TIE as a function of β. Its minima exhibit a non-monotonic trend. (C) Reconstruction objective as a function of β. (D) Conditional independence loss as a function of β. (E) Traversal of latent encoding in bottleneck neurons for small and large β. One neuron is held fixed while the other is modulated to generate reconstructions. We can see worsening reconstruction with higher β, while units in the bottleneck encode for seemingly orthogonal axes of motion.



Chapter IV.

Discussion

The research outlined in this thesis explored the notion of disentangled representations, a core idea in both artificial and biological learning systems. Specifically, disentangled representations were presented through a probabilistic interpretation, and examined in the context of extracting statistically independent latent variables from the β-VAE architecture. General theorems in Variational Bayesian inference were proven in the context of the β-VAE architecture, and an analytically tractable model was introduced. Simulations with deep β-VAE models on real world synthetic datasets with various source types and configurations confirmed insights from the analytical model and theorems. With the goal of disentangling independent latent factors, β-VAE may lead to suboptimal values of β, as it is enforcing conditional independence of its bottleneck representation.

Disentangling Representations is an exciting area of research, and there are other perspectives on the definition as alluded throughout this text (Bengio et al., 2013; Burgess et al., 2018). Many of these definitions do not center on statistical formalism, and instead take into account symmetry transformations and the manifold structure in data (Bengio et al., 2013; DiCarlo & Cox, 2007; Higgins et al., 2018), or focus on disentangling using adversarial approaches (Denton et al., 2017; Tran, Yin, & Liu, 2017; John, Mou, Bahuleyan, & Vechtomova, 2018). The topic has also been studied in



supervised contexts (Siddharth et al., 2017). The research taking place in the general overall direction of disentangling representations is important to downstream computer vision and natural language tasks, and may prove useful in a variety of applications. The study of these independent representations may also provide a useful model with which to study disentangling in biological contexts. Better understanding disentangled representations also is related to the emerging field of interpretability and the development of new data analysis tasks. All together, the work presented in this paper represents a useful exploration and perspective into a field that promises better understanding and utility of both biological and artificial systems, ranging from theory to real world applications.



Appendix 1.

Integrating out data from the β-VAE Objective

Averaging the β-VAE Objective with respect to the data distribution p(x), we arrive at:

$$L(\boldsymbol{\theta}, \boldsymbol{\phi}; \beta) \equiv \mathbb{E}_{p(\mathbf{x})}\left[\mathcal{L}(\boldsymbol{\theta}, \boldsymbol{\phi}; \beta)\right]$$
$$= \mathbb{E}_{p(\mathbf{x})}\left[\mathbb{E}_{q_{\boldsymbol{\phi}}(\mathbf{z}|\mathbf{x})}\left[\log p_{\boldsymbol{\theta}}(\mathbf{x}|\mathbf{z})\right] - \beta D_{KL}\left(q_{\boldsymbol{\phi}}(\mathbf{z}|\mathbf{x})\|p(\mathbf{z})\right)\right]. \tag{55}$$

Using the reparameterization trick to write $\mathbf{z} = \mu_{\mathbf{z}} + \boldsymbol{\Sigma}_{\mathbf{z}}^{\mathbf{1/2}}\epsilon$ with $\epsilon \sim (\mathbf{0}, \mathbf{I}_k)$ in $\mathbb{E}_{q_{\phi}(\mathbf{z}|\mathbf{x})}\left[\log p_{\theta}(\mathbf{x}|\mathbf{z})\right]$. Subsequently,

$$\mathbb{E}_{\mathbf{z} \sim q_{\boldsymbol{\phi}}(\mathbf{z}|\mathbf{x})}\left[\log p_{\boldsymbol{\theta}}(\mathbf{x}|\mathbf{z})\right]$$
$$= \mathbb{E}_{\mathbf{z} \sim q_{\boldsymbol{\phi}}(\mathbf{z}|\mathbf{x})}\left[\log \mathcal{N}(\mathbf{x}; \mathbf{Dz} + \mathbf{b}^D, \mathbf{I}_N)\right]$$
$$= \mathbb{E}_{\boldsymbol{\epsilon} \sim \mathcal{N}(0,1)}\left[\log \mathcal{N}(\mathbf{x}; \mathbf{D}(\boldsymbol{\mu}_{\mathbf{z}} + \boldsymbol{\Sigma}_{\mathbf{z}}^{\mathbf{1/2}}\boldsymbol{\epsilon}) + \mathbf{b}^D, \mathbf{I}_N)\right]$$
$$= -\frac{N}{2}\log(2\pi) - \frac{1}{2}(\mathbf{D}\boldsymbol{\mu}_{\mathbf{z}} + \mathbf{b}^D - \mathbf{x})^2$$
$$\quad - \frac{1}{2}\mathbb{E}_{\boldsymbol{\epsilon} \sim \mathcal{N}(0,1)}\left[\boldsymbol{\epsilon}^{\top}(\mathbf{D}\boldsymbol{\Sigma}_{\mathbf{z}}^{1/2})^{\top}(\mathbf{D}\boldsymbol{\Sigma}_{\mathbf{z}}^{1/2})\boldsymbol{\epsilon}\right].$$

Introducing a source term J allows for the use of a trick to calculate the final term above.



$$Z[\mathbf{J}] = \int \frac{d\mathbf{z}}{(2\pi)^{n/2}\sqrt{\det \boldsymbol{\Sigma_z}}} \exp\left(-\frac{1}{2}\mathbf{z}^\top \boldsymbol{\Sigma_z}^{-1}\mathbf{z} + \mathbf{J}^\top \mathbf{A}\mathbf{z}\right), \qquad (61)$$

Now, we can differentiate with respect to the source

$$\left(\frac{\delta}{\delta \mathbf{J}}\right)^\top \left(\frac{\delta}{\delta \mathbf{J}}\right) Z[\mathbf{J}]\bigg|_{\mathbf{J}=0} = \mathbb{E}_{\mathbf{z}\sim\mathcal{N}(0,\boldsymbol{\Sigma_z})}\left[\mathbf{z}^\top \mathbf{A}^\top \mathbf{A}\mathbf{z}\right] \qquad (62)$$

Now, using the Gaussian integral in $Z[\mathbf{J}]$ we obtain,

$$Z[\mathbf{J}] = \exp\left\{\frac{1}{2}(\mathbf{J}\mathbf{A})^\top \boldsymbol{\Sigma_z}(\mathbf{J}\mathbf{A})\right\}. \qquad (63)$$

and arrive at

$$\mathbb{E}_{\mathbf{z}\sim\mathcal{N}(0,\boldsymbol{\Sigma_z})}\left[\mathbf{z}^\top \mathbf{A}^\top \mathbf{A}\mathbf{z}\right] = \mathrm{Tr}(\mathbf{A}\boldsymbol{\Sigma_z}\mathbf{A}^\top) \qquad (64)$$

Using the above equation, we can re-examine the reconstruction objective and also calculate the conditional independence loss:

$$\mathbb{E}_{\mathbf{z}\sim q_{\boldsymbol{\phi}}(\mathbf{z}|\mathbf{x})}[\log p_{\boldsymbol{\theta}}(\mathbf{x}|\mathbf{z})] = -(\mathbf{D}\boldsymbol{\mu_z} + \mathbf{b}^D - \mathbf{x})^2$$
$$- \mathrm{Tr}(\mathbf{D}^\top \mathbf{D}\boldsymbol{\Sigma_z}). \qquad (65)$$

$$D_{KL}(q_{\boldsymbol{\phi}}(\mathbf{z}|\mathbf{x})\|p(\mathbf{z}))$$
$$= -\frac{1}{2}\left(k + \mathrm{Tr}\log\boldsymbol{\Sigma_z} - \boldsymbol{\mu_z}^\top \boldsymbol{\mu_z} - \mathrm{Tr}\boldsymbol{\Sigma_z}\right). \qquad (66)$$

Taken together, the objective function to maximize without constants is the following.

$$L(\boldsymbol{\theta},\boldsymbol{\phi};\beta) = \frac{1}{2}\mathbb{E}_{p(\mathbf{x})}\left[-(\mathbf{D}\boldsymbol{\mu_z} + \mathbf{b}^D - \mathbf{x})^\top(\mathbf{D}\boldsymbol{\mu_z} + \mathbf{b}^D - \mathbf{x})\right.$$
$$\left. - \beta\boldsymbol{\mu_z}^\top \boldsymbol{\mu_z} - \mathrm{Tr}(\mathbf{D}^\top \mathbf{D}\boldsymbol{\Sigma_z}) + \beta\mathrm{Tr}\log\boldsymbol{\Sigma_z} - \beta\mathrm{Tr}\boldsymbol{\Sigma_z}\right]. \qquad (67)$$

After plugging in the definition of $\boldsymbol{\mu_z}, \boldsymbol{\Sigma_z}$ and performing Gaussian integrals in $\mathbf{x}$, we arrive at equation 35 presented in the results section.



Appendix 2.

Deriving the Ground Truth Posterior

Since $\eta$ and $\mathbf{s}$ are independently normally distributed in our generative process, $\mathbf{s}$ and $\eta$ are jointly normal. But since $p(\mathbf{s}, \eta)$ is $p(\mathbf{s}, \mathbf{x})$ up to a coordinate transformation, $p(\mathbf{s}, \mathbf{x})$ is also normal. $\mathbf{s} \in \mathbb{R}^{\mathbf{k}}$, $\mathbf{x} \in \mathbb{R}^{\mathbf{N}}$, $(\mathbf{s}, \mathbf{x}) \in \mathbb{R}^{\mathbf{N+k}}$.

Accordingly, $\mathbf{s}$ and $\mathbf{x}$ can be considered as partitioning a $(N+k)$-dimensional normal distribution $p((\mathbf{s}, \mathbf{x}))$. So finding $p_{\text{g-t}}(\mathbf{s}|\mathbf{x})$, involves using the formula for conditioning multivariate normal distribution:

$$p_{\text{g-t}}(\mathbf{s}|\mathbf{x}) = \mathcal{N}(\boldsymbol{\mu}_{\mathbf{s}|\mathbf{x}}, \boldsymbol{\Sigma}_{\mathbf{s}|\mathbf{x}}),$$

where

$$\boldsymbol{\mu}_{\mathbf{s}|\mathbf{x}} = \boldsymbol{\mu}_{\mathbf{s}} + \text{Cov}(\mathbf{x}, \mathbf{s})^{\top} (\boldsymbol{\Sigma}_{\mathbf{x}})^{-1} (\mathbf{x} - \boldsymbol{\mu}_{\mathbf{x}})$$
$$\boldsymbol{\Sigma}_{\mathbf{s}|\mathbf{x}} = \boldsymbol{\Sigma}_{\mathbf{s}} - \text{Cov}(\mathbf{x}, \mathbf{s})^{\top} (\boldsymbol{\Sigma}_{\mathbf{x}})^{-1} \text{Cov}(\mathbf{x}, \mathbf{s}).$$

Specializing to the generative case described earlier,

$$\text{Cov}(\mathbf{x}, \mathbf{s}) = \text{Cov}(\mathbf{A}\mathbf{s} + \boldsymbol{\eta}, \mathbf{s}) = \mathbf{A}\boldsymbol{\Sigma}_{\mathbf{s}} + \text{Cov}(\mathbf{s}, \boldsymbol{\eta}) = \mathbf{A}.$$

$\boldsymbol{\Sigma}_{\mathbf{x}} = \mathbf{A}\mathbf{A}^{\top} + \mathbf{I}_N$. Thus

$$\boldsymbol{\mu}_{\mathbf{s}|\mathbf{x}} = \mathbf{A}^{\top}(\mathbf{A}\mathbf{A}^{\top} + \mathbf{I}_N)^{-1}\mathbf{x} = (\mathbf{A}^{\top}\mathbf{A} + \mathbf{I}_k)^{-1}\mathbf{A}^{\top}\mathbf{x},$$



The second equality above uses the *matrix push-through identity*: For any matrices $\mathbf{U} \in \mathbb{R}^{N \times k}, \mathbf{V} \in \mathbb{R}^{k \times N}$,

$$(\mathbf{I}_N + \mathbf{U}\mathbf{V})^{-1}\mathbf{U} = \mathbf{U}(\mathbf{I}_k + \mathbf{V}\mathbf{U})^{-1}. \tag{67}$$

Now the covariance,

$$\begin{aligned}
\boldsymbol{\Sigma}_{\mathbf{s}|\mathbf{x}} &= \mathbf{I}_k - \mathbf{A}^\top(\mathbf{A}\mathbf{A}^\top + \mathbf{I}_N)^{-1}\mathbf{A} \\
&= \mathbf{I}_k - \mathbf{A}^\top\mathbf{A}(\mathbf{A}^\top\mathbf{A} + \mathbf{I}_k)^{-1} \\
&= \mathbf{I}_k - \mathbf{A}^\top\mathbf{A}[(\mathbf{A}^\top\mathbf{A})^{-1} - (\mathbf{A}^\top\mathbf{A})^{-1}(\mathbf{A}^\top\mathbf{A} + \mathbf{I}_k)^{-1}] \\
&= (\mathbf{A}^\top\mathbf{A} + \mathbf{I}_k)^{-1},
\end{aligned} \tag{68}$$

The third equality uses the *Woodbury matrix identity*, which says that for any invertible matrix $\mathbf{B} \in \mathbb{R}^{N \times N}$ and size compatible matrices $\mathbf{U} \in \mathbb{R}^{N \times k}$ and $\mathbf{V} \in \mathbb{R}^{k \times N}$:

$$(\mathbf{B} + \mathbf{U}\mathbf{V})^{-1} = \mathbf{B}^{-1} - \mathbf{B}^{-1}\mathbf{U}(\mathbf{I}_k + \mathbf{V}\mathbf{B}^{-1}\mathbf{U})^{-1}\mathbf{V}\mathbf{B}^{-1}. \tag{69}$$





## Appendix 3.

## Proof of Proposition 2

Consider an objective function:

$$O(\boldsymbol{\kappa}; \beta) = A(\boldsymbol{\kappa}) - \beta B(\boldsymbol{\kappa}), \tag{70}$$

The function is given by the sum of two terms, to be maximized over $\boldsymbol{\kappa}$.

$$\boldsymbol{\kappa}^*(\beta) = \arg\max_{\boldsymbol{\kappa}} O(\boldsymbol{\kappa}, \beta) \tag{71}$$

As $\beta$ increases $B(\kappa^*(\beta))$ is nonincreasing.
The proof uses contradiction. Let $\beta_2 > \beta_1$ and

$$\boldsymbol{\kappa}_1 \equiv \boldsymbol{\kappa}^*(\beta_1), \qquad \boldsymbol{\kappa}_2 \equiv \boldsymbol{\kappa}^*(\beta_2). \tag{72}$$

Which leads to the following statement. Note that the the first line is an identity, and the second line follows from the optimality of $\boldsymbol{\kappa}_2$ at $\beta = \beta_2$.

$$\begin{aligned} O(\boldsymbol{\kappa}_1, \beta_1) &= O(\boldsymbol{\kappa}_1, \beta_2) + (\beta_2 - \beta_1)B(\boldsymbol{\kappa}_1) \\ &\leq O(\boldsymbol{\kappa}_2, \beta_2) + (\beta_2 - \beta_1)B(\boldsymbol{\kappa}_1), \end{aligned} \tag{73}$$

This leads to a contradiction, assuming $B(\boldsymbol{\kappa}_2) > B(\boldsymbol{\kappa}_1)$.

$$\begin{aligned} O(\boldsymbol{\kappa}_2, \beta_2) &+ (\beta_2 - \beta_1)B(\boldsymbol{\kappa}_1) \\ &< O(\boldsymbol{\kappa}_2, \beta_2) + (\beta_2 - \beta_1)B(\boldsymbol{\kappa}_2) = O(\boldsymbol{\kappa}_2, \beta_1). \end{aligned} \tag{74}$$

This follows from the equality provided at the beginning of this proof. Combined with equation 58, this implies

$$O(\boldsymbol{\kappa}_1, \beta_1) < O(\boldsymbol{\kappa}_2, \beta_1) \tag{75}$$

Thus, if $\beta_1 < \beta_2$, then $B(\kappa_2) \leq B(\kappa_1)$



Appendix 4.

MIE/TIE Derivation

If we let,

$$\boldsymbol{\mu}_{\mathbf{z}|\mathbf{x}} \equiv \mathbf{F}\mathbf{x}, \qquad \boldsymbol{\Sigma}_{\mathbf{z}|\mathbf{x}} \equiv \mathbf{E}. \tag{76}$$

MIE can be written as

$$\begin{aligned}
\text{MIE} =& \mathbb{E}_{p(\mathbf{x})}\big[D_{KL}(q_{\boldsymbol{\phi}}(\mathbf{z}|\mathbf{x})\|p_{\boldsymbol{\theta}}(\mathbf{x}|\mathbf{z}))\big] \\
=& \frac{1}{2}\mathbb{E}_{p(\mathbf{x})}\bigg[(\mathbf{F}\mathbf{x} - \boldsymbol{\mu}_{\mathbf{z}})^{\top}\mathbf{E}^{-1}(\mathbf{F}\mathbf{x} - \boldsymbol{\mu}_{\mathbf{z}}) \\
& + \text{Tr}(\mathbf{E}^{-1}\boldsymbol{\Sigma}_{\mathbf{z}}) - \log\left(\frac{\det \boldsymbol{\Sigma}_{\mathbf{z}}}{\det \mathbf{E}}\right) - k\bigg].
\end{aligned} \tag{77}$$

After plugging in the definition of $\boldsymbol{\mu}_{\mathbf{z}}, \boldsymbol{\Sigma}_{\mathbf{z}}$ and performing Gaussian integrals in $\mathbf{x}$, we arrive at the MIE/TIE equation.

At network optimum, $p_{\boldsymbol{\theta}}(\mathbf{z}|\mathbf{x})$ is equal to $p(\mathbf{s}|\mathbf{x})$ when $\mathbf{D}$ to $\mathbf{A}$. Accordingly replacing $\mathbf{D}$ with $\mathbf{A}$ yields the TIE.



# Appendix 5.

## Model Posterior Derivation

To calculate the model posterior using Bayes Rule, we need to calculate the *evidence* $p_{\boldsymbol{\theta}}(\mathbf{x})$

$$p_{\boldsymbol{\theta}}(\mathbf{x}) = \int_{\mathbb{R}^k} d\mathbf{z} \, p_{\boldsymbol{\theta}}(\mathbf{x}|\mathbf{z}) p(\mathbf{z}) \tag{78}$$

$$= \int_{\mathbb{R}^k} d\mathbf{z} \, \mathcal{N}(\mathbf{Dz}, \mathbf{I}_N) \mathcal{N}(\mathbf{0}, \mathbf{I}_k) \tag{79}$$

$$= \mathcal{N}(\mathbf{0}, (\mathbf{DD}^\top + \mathbf{I}_N)), \tag{80}$$

Using the woodbury matrix equation and some others from the ground truth posterior derivation,

$$p_{\boldsymbol{\theta}}(\mathbf{x}|\mathbf{z}) = \mathcal{N}((\mathbf{DD}^\top + \mathbf{I}_N)^{-1}\mathbf{D}^\top, (\mathbf{DD}^\top + \mathbf{I}_N)^{-1})$$
$$\equiv \mathcal{N}(\boldsymbol{\mu}_{\mathbf{z}|\mathbf{x}}, \boldsymbol{\Sigma}_{\mathbf{z}|\mathbf{x}}). \tag{81}$$



## Appendix 6.

### Table of Useful Expressions

| Term | Mathematical Expression |
|------|------------------------|
| Prior | $p(\mathbf{z})$ |
| Model Posterior | $p_{\boldsymbol{\theta}}(\mathbf{z}|\mathbf{x})$ |
| Ground-Truth Posterior | $p_{\text{g-t}}(\mathbf{z}|\mathbf{x})$ |
| Inference Model | $q_{\boldsymbol{\phi}}(\mathbf{z}|\mathbf{x})$ |
| Data Log-Likelihood | $\log p_{\boldsymbol{\theta}}(\mathbf{x})$ |
| Reconstruction Objective | $\mathbb{E}_{q_{\boldsymbol{\phi}}(\mathbf{z}|\mathbf{x})}\left[\log p_{\boldsymbol{\theta}}(\mathbf{x}|\mathbf{z})\right]$ |
| Conditional Independence Loss | $D_{KL}\left(q_{\boldsymbol{\phi}}(\mathbf{z}|\mathbf{x})\|p(\mathbf{z})\right)$ |
| MIE | $\mathbb{E}_{p(\mathbf{x})}[D_{KL}(q_{\boldsymbol{\phi}}(\mathbf{z}|\mathbf{x})\|p_{\theta}(\mathbf{z}|\mathbf{x}))]$ |
| TIE | $\mathbb{E}_{p(\mathbf{x})}[D_{KL}(q_{\boldsymbol{\phi}}(\mathbf{z}|\mathbf{x})\|p_{\text{g-t}}(\mathbf{z}|\mathbf{x}))]$ |
| Evidence Lower Bound (ELBO) | $\mathbb{E}_{q_{\boldsymbol{\phi}}(\mathbf{z}|\mathbf{x})}\left[\log p_{\boldsymbol{\theta}}(\mathbf{x}|\mathbf{z})\right]$ $-D_{KL}\left(q_{\boldsymbol{\phi}}(\mathbf{z}|\mathbf{x})\|p(\mathbf{z})\right)$ |



References


Alemi, A., Poole, B., Fischer, I., Dillon, J., Saurus, R. A., & Murphy, K. (2018). An information-theoretic analysis of deep latent-variable models.

Ba, J., Mnih, V., & Kavukcuoglu, K. (2014). Multiple object recognition with visual attention. arXiv preprint arXiv:1412.7755.

Baldi, P. (2012, June). Autoencoders, unsupervised learning, and deep architectures. In Proceedings of ICML workshop on unsupervised and transfer learning (pp. 37-49).

Bengio, Y., Courville, A., & Vincent, P. (2013). Representation learning: A review and new perspectives. IEEE transactions on pattern analysis and machine intelligence, 35 (8), 1798–1828.

Burgess, C. P., Higgins, I., Pal, A., Matthey, L., Watters, N., Desjardins, G., & Lerchner, A. (2018). Understanding disentangling in β-vae. arXiv preprint arXiv:1804.03599 .

Cadieu, C. F., Hong, H., Yamins, D. L., Pinto, N., Ardila, D., Solomon, E. A., ... & DiCarlo, J. J. (2014). Deep neural networks rival the representation of primate IT cortex for core visual object recognition. PLoS computational biology, 10(12), e1003963.

Cichy, R. M., Pantazis, D., & Oliva, A. (2014). Resolving human object recognition in space and time. Nature neuroscience, 17(3), 455.

Dayan, P., Abbott, L. F., et al. (2001). Theoretical neuroscience (Vol. 806). Cambridge, MA: MIT Press.

Denton, E. L. (2017). Unsupervised learning of disentangled representations from video. In Advances in neural information processing systems (pp. 4414-4423).

DiCarlo, J. J., Zoccolan, D., & Rust, N. C. (2012). How does the brain solve visual object recognition?. Neuron, 73(3), 415-434.

Doersch, C. (2016). Tutorial on variational autoencoders. arXiv preprint arXiv:1606.05908 .

Douglas, R. J., & Martin, K. A. (1991). A functional microcircuit for cat visual cortex. The Journal of Physiology, 440(1), 735-769.





Fusi, S., Miller, E. K., & Rigotti, M. (2016). Why neurons mix: high dimensionality for higher cognition. Current opinion in neurobiology, 37, 66-74.

Geirhos, R., Janssen, D. H., Schütt, H. H., Rauber, J., Bethge, M., & Wichmann, F. A. (2017). Comparing deep neural networks against humans: object recognition when the signal gets weaker. arXiv preprint arXiv:1706.06969.

Géron, A. (2017). Hands-on machine learning with Scikit-Learn and TensorFlow: concepts, tools, and techniques to build intelligent systems. " O'Reilly Media, Inc.".

Goodfellow, I., Bengio, Y., & Courville, A. (2016). Deep learning. MIT press.

Gurtubay-Antolin, A., Rodriguez-Herreros, B., & Rodríguez-Fornells, A. (2015). The speed of object recognition from a haptic glance: event-related potential evidence. Journal of neurophysiology, 113(9), 3069-3075.

Hassabis, D., Kumaran, D., Summerfield, C., & Botvinick, M. (2017). Neuroscience-inspired artificial intelligence. Neuron, 95(2), 245-258.

Higgins, I., Amos, D., Pfau, D., Racaniere, S., Matthey, L., Rezende, D., & Lerchner, A. (2018). Towards a Definition of Disentangled Representations. arXiv preprint arXiv:1812.02230.

Higgins, I., Matthey, L., Pal, A., Burgess, C., Glorot, X., Botvinick, M., . . . Lerchner, A. (2017). beta-vae: Learning basic visual concepts with a constrained variational framework. ICLR, 2 (5), 6.

Hinton, G. E., & Zemel, R. S. (1994). Autoencoders, minimum description length and Helmholtz free energy. In Advances in neural information processing systems (pp. 3-10).

Huang, X., Liu, M.-Y., Belongie, S., & Kautz, J. (2018, September). Multimodal unsupervised image-to-image translation. In The european conference on computer vision (eccv).

Hyvarinen, A., & Oja, E. (2000). Independent component analysis: algorithms and applications. Neural networks, 13 (4-5), 411–430.

John, V., Mou, L., Bahuleyan, H., & Vechtomova, O. (2018). Disentangled representation learning for text style transfer. arXiv preprint arXiv:1808.04339 .

Jones, Eric, Travis Oliphant, and Pearu Peterson. "{SciPy}: Open source scientific tools for {Python}." (2014).





Jordan, M. I., Ghahramani, Z., Jaakkola, T. S., & Saul, L. K. (1999). An introduction to variational methods for graphical models. Machine learning, 37 (2), 183–233.

Karras, T., Laine, S., & Aila, T. (2019). A style-based generator architecture for generative adversarial networks. In Proceedings of the ieee conference on computer vision and pattern recognition (pp. 4401–4410)

Khemakhem, I., Kingma, D. P., & Hyvarinen, A. (2019). Variational autoencoders and nonlinear ica: A unifying framework. arXiv preprint arXiv:1907.04809 .

Kingma, D. P., & Welling, M. (2013). Auto-encoding variational bayes. arXiv preprint arXiv:1312.6114 .

Kingma, D. P., & Ba, J. (2014). Adam: A method for stochastic optimization. arXiv preprint arXiv:1412.6980.

Kingma, D. P., & Welling, M. (2019). An introduction to variational autoencoders. arXiv preprint arXiv:1906.02691 .

Koster-Hale, J., & Saxe, R. (2013). Theory of mind: a neural prediction problem. Neuron, 79(5), 836-848.

Kravitz, D. J., Saleem, K. S., Baker, C. I., Ungerleider, L. G., & Mishkin, M. (2013). The ventral visual pathway: an expanded neural framework for the processing of object quality. Trends in cognitive sciences, 17(1), 26-49.

Lample, G., Zeghidour, N., Usunier, N., Bordes, A., Denoyer, L., & Ranzato, M. (2017). Fader networks: Manipulating images by sliding attributes. In Advances in neural information processing systems (pp. 5967–5976).

Li, N., Cox, D. D., Zoccolan, D., & DiCarlo, J. J. (2009). What response properties do individual neurons need to underlie position and clutter "invariant" object recognition?. Journal of Neurophysiology, 102(1), 360-376.

LeCun, Y., Bengio, Y., & Hinton, G. (2015). Deep learning. nature, 521(7553), 436.

Liu, W., Wang, Z., Liu, X., Zeng, N., Liu, Y., & Alsaadi, F. E. (2017). A survey of deep neural network architectures and their applications. Neurocomputing, 234, 11-26.

Locatello, F., Bauer, S., Lucic, M., Gelly, S., Scholkopf, B., & Bachem, O. (2018). Challenging common assumptions in the unsupervised learning of disentangled representations. arXiv preprint arXiv:1811.12359 .





Lotfi, E., Araabi, B. N., Ahmadabadi, M. N., & Schwabe, L. (2014, November). Biological constrained learning of parameters in a recurrent neural network-based model of the primary visual cortex. In 2014 21th Iranian Conference on Biomedical Engineering (ICBME) (pp. 292-297). IEEE.

Pedregosa, F., Varoquaux, G., Gramfort, A., Michel, V., Thirion, B., Grisel, O., ... & Vanderplas, J. (2011). Scikit-learn: Machine learning in Python. Journal of machine learning research, 12(Oct), 2825-2830.

Pinto, N., Barhomi, Y., Cox, D. D., & DiCarlo, J. J. (2011, January). Comparing state-of-the-art visual features on invariant object recognition tasks. In 2011 IEEE workshop on Applications of computer vision (WACV) (pp. 463-470). IEEE.

Pinto, N., Cox, D. D., Corda, B., Doukhan, D., & DiCarlo, J. J. (2008). Why is real-world object recognition hard? Establishing honest benchmarks and baselines for object recognition. Proceedings of the COSYNE, 8.

Oliphant, T. E. (2007). Python for scientific computing. Computing in Science & Engineering, 9(3), 10-20.

Rolls, E. T. (2012). Invariant visual object and face recognition: neural and computational bases, and a model, VisNet. Frontiers in computational neuroscience, 6, 35.

Rosenblatt, F. (1958). The perceptron: a probabilistic model for information storage and organization in the brain. Psychological review, 65(6), 386.

Sabour, S., Frosst, N., & Hinton, G. E. (2017). Dynamic routing between capsules. In Advances in neural information processing systems (pp. 3856-3866).

Sikka, H., Zhong, W., Yin, J., & Pehlevan, C. (2019). A Closer Look at Disentangling in β-VAE. *arXiv preprint arXiv:1912.05127.*

Sanner, M. F. (1999). Python: a programming language for software integration and development. J Mol Graph Model, 17(1), 57-61.

Siddharth, N., Paige, B., Van de Meent, J.-W., Desmaison, A., Goodman, N., Kohli, P., . . . Torr, P. (2017). Learning disentangled representations with semi-supervised deep generative models. In Advances in neural information processing systems (pp. 5925–5935).

Tipping, M. E., & Bishop, C. M. (1999). Probabilistic principal component analysis. Journal of the Royal Statistical Society: Series B (Statistical Methodology), 61 (3), 611–622.





Tran, L., Yin, X., & Liu, X. (2017). Disentangled representation learning gan for pose-invariant face recognition. In Proceedings of the ieee conference on computer vision and pattern recognition (pp. 1415–1424).

Van Der Walt, S., Colbert, S. C., & Varoquaux, G. (2011). The NumPy array: a structure for efficient numerical computation. Computing in Science & Engineering, 13(2), 22.

Van Essen, D. C., Anderson, C. H., & Felleman, D. J. (1992). Information processing in the primate visual system: an integrated systems perspective. Science, 255(5043), 419-423.

Yamins, D. L., Hong, H., Cadieu, C. F., Solomon, E. A., Seibert, D., & DiCarlo, J. J. (2014). Performance-optimized hierarchical models predict neural responses in higher visual cortex. Proceedings of the National Academy of Sciences, 111(23), 8619-8624.

Yamins, D. L., & DiCarlo, J. J. (2016). Using goal-driven deep learning models to understand sensory cortex. Nature neuroscience, 19(3), 356.

Ullman, S. (2000). High-level vision: Object recognition and visual cognition. MIT press.